%% file: neurips_2025.tex
\documentclass{article}

\input{math_commands}

\usepackage[preprint]{neurips_2025}

\usepackage[utf8]{inputenc} 
\usepackage[T1]{fontenc}    
\usepackage{hyperref}       
\usepackage{url}            
\usepackage{booktabs}       
\usepackage{amsfonts}       
\usepackage{nicefrac}       
\usepackage{microtype}      

\usepackage[textsize=tiny]{todonotes}

\usepackage{wrapfig}

\usepackage[utf8]{inputenc} 
\usepackage[T1]{fontenc}    
\usepackage{hyperref}       
\usepackage{url}            
\usepackage{booktabs}       
\usepackage{amsfonts}       
\usepackage{nicefrac}       
\usepackage{microtype}      
\usepackage{xspace}
\usepackage[inline]{enumitem}
\usepackage{natbib}
\usepackage{amsmath,amsfonts,amssymb}
\usepackage{pgf}
\usepackage{wrapfig}
\usepackage{amssymb}
\usepackage{amsthm}
\usepackage{multirow}
\usepackage{caption}
\usepackage{wrapfig}

\usepackage{xcolor}
\usepackage{xspace}

\usepackage{wrapfig}  
\usepackage{lipsum}   

\newcommand{\parbold}[1]{\textbf{#1.}\xspace}
\newcommand{\Gtr}{\gG_\mathrm{tr}}
\newcommand{\Gval}{\gG_\mathrm{val}}
\newcommand{\Gtest}{\gG_\mathrm{test}}

\newcommand{\Vattack}{\gV_\mathrm{att}}

\newcommand{\method}{EvA\xspace}
\newcommand{\mathfloor}[1]{\lfloor #1 \rfloor}

\newcommand{\pubmed}{\texttt{Pubmed}\xspace}
\newcommand{\coraml}{\texttt{CoraML}\xspace}
\newcommand{\citeseer}{\texttt{Citeseer}\xspace}
\newcommand{\photo}{\texttt{Amazon-Photo}\xspace}
\newcommand{\computers}{\texttt{Amazon-Computers}\xspace}
\newcommand{\arxiv}{\texttt{Ogbn-Arxiv}\xspace}

\def\vepsilon{\bm{\epsilon}}

\title{EvA: Evolutionary Attacks on Graphs}

\author{%
  Sadegh Akhondzadeh\thanks{Equal Contributions } \ $^{1}$ \quad Soroush H. Zargarbashi$^{* 2}$ \quad Jimin Cao$^{1}$ \quad
  Aleksandar Bojchevski$^{1}$ \\
  $^{1}$ University of Cologne, $^{2}$ CISPA Helmholtz Center for Information Security  \\
  \texttt{[akhondzadeh,zargarbashi, jcao, bojchevski]@cs.uni-koeln.de} \\
}

\begin{document}

\maketitle

\begin{abstract}
Even a slight perturbation in the graph structure can cause a significant drop in the accuracy of graph neural networks (GNNs). Most existing attacks leverage gradient information to perturb edges. This relaxes the attack's optimization problem from a discrete to a continuous space, resulting in solutions far from optimal. It also restricts the adaptability of the attack to non-differentiable objectives. Instead, we introduce a few simple yet effective enhancements of an evolutionary-based algorithm to solve the discrete optimization problem directly. Our \underline{Ev}olutionary \underline{A}ttack (\method) works with any black-box model and objective, eliminating the need for a differentiable proxy loss. This allows us to design two novel attacks that reduce the effectiveness of robustness certificates and break conformal sets.  The memory complexity of our attack is linear in the attack budget. 
Among our experiments, \method shows $\sim$11\% additional drop in accuracy on average compared to the best previous attack, 
revealing significant untapped potential in designing attacks. \looseness=-1
\end{abstract}

\section{Introduction}
\label{sec:intro}
Given the widespread applications of graph neural networks (GNNs), it's crucial to study their robustness to natural and adversarial noise.
In node classification, GNNs leverage the edge information to improve their performance. 
However, 
adding or removing a few edges can drastically decrease their accuracy, even below the performance of an MLP that ignores the graph structure entirely.
The vast majority of adversarial attacks on the graph structure are gradient-based. 

\begin{figure}
    \centering
\includegraphics[width=0.8\linewidth]{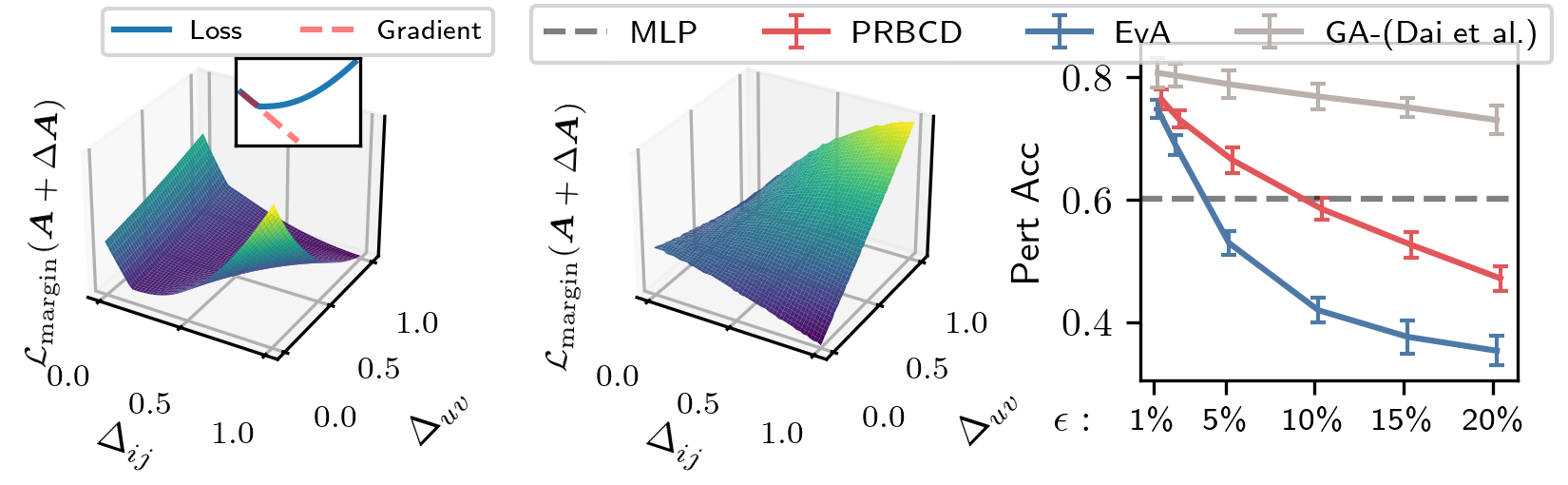}
\caption{
We compute $\gL_\mathrm{margin}(\mA + \Delta \mA)$ where $\Delta \mA = \ve_i \ve_j^\top \Delta_{ij} + \ve_u \ve_v^\top \Delta_{uv}$ and $\ve_i$ is the $i$-the cannonical vector.
[Left] The loss landscape is non-linear, and the gradient does not always indicate the loss direction when we flip an edge (e.g. gradient suggests decrease, but loss increases). [Middle] Due to non-convexity, the effect of flipping each edge separately (e.g. loss decreases) can differ from flipping both edges simultaneously (e.g. loss increases). This happens for many edges (\autoref{app: issues_grad}).  [Right] \method does not suffer from this issue and outperforms both PRBCD and search-based attacks.}
    \label{fig: motivation_figure}
\end{figure}

However, gradient-based attacks face several challenges in this setting: \begin{enumerate*}[label=(\roman*)]
    \item To tackle the original discrete combinatorial optimization problem we have to relax the domain from $\{0, 1\}$ to $[0, 1]$;
    \item The gradients only provides local information and cannot accurately reflect the actual loss landscape when edges are flipped (see \autoref{fig: motivation_figure});
    \item Similarly, the gradient only reflects the effect of flipping \emph{a single} edge at a time, but the effect on the loss can be different (even opposite) when two or more edges are flipped simultaneously (see \autoref{fig: motivation_figure});
    \item We need a differentiable proxy loss function since the original attack objective is often not differentiable (e.g. accuracy). A common choice is cross-entropy which is suboptimal as a proxy \citep{geisler2023robustnessgraphneuralnetworks};
    \item White-box access to the model is necessary, which limits the applicability or requires surrogate models.;
    \item Defense against such attacks might carry a false sense of security by only  obfuscating gradients \citep{athalye2018obfuscated, geisler2023robustnessgraphneuralnetworks};
    \item Although the adjacency matrix is often sparse, the gradients w.r.t. it are not. Therefore, the memory complexity of these attacks grows quadratically w.r.t the number of nodes, for which tricks like block coordinate descent are needed \citep{prbcd}.
    \end{enumerate*}
    
These challenges suggest that we should try to directly solve the original (combinatorial discrete) optimization problem, and not to rely on differentiation.
A natural alternative is search. 
Indeed, \citet{dai2018adversarial} implemented a baseline genetic-based search for attacking the edges. However, their approach was not competitive with gradient-based attacks, largely due to poor design choices in the loss function and mutation strategies.
While search-based attacks have been promptly forgotten since, we show that by carefully designing the components of a meta-heuristic pipeline we can outperform state-of-the-art gradient-based attacks by a significant margin. As shown in \autoref{fig: motivation_figure}[Right], \method not only outperforms the previous method based on Genetic Algorithm, but also outperforms PRBCD, the previous state-of-the-art, by a large margin.
Our model-agnostic \underline{ev}olutionary \underline{a}ttack (\method) 
explores the space of possible perturbations with a genetic algorithm (GA) without gradient information. We avoid domain relaxation by directly minimizing the (non-differentiable) accuracy over the space of binary matrices.
Our attack easily extends to other objectives. We show this by
defining two novel attacks on graphs that break conformal guarantees or reduce the effectiveness of robustness certificates (see \autoref{sec:local-and-targetted}). Importantly, this extension is automatic -- we only need black-box access to the objective, making our attack adaptive \cite{mujkanovic2023defensesgraphneuralnetworks}. In contrast, gradient-based attacks for these new objectives require substantial additional effort (e.g. to tailor the right relaxations).

Our implementation requires $\gO(\epsilon \cdot E)$ memory complexity where $\epsilon$ is the perturbation budget, and $E$ is the number of edges. Given more time or more memory we can increase our performance due to the open-ended nature of the search. This is a missing characteristic in SOTA attacks such as PRBCD \cite{prbcd}. For larger graphs where the search space is considerably larger, we apply a divide and conquer strategy that improves both PRBCD and \method, with \method still outperforming.

In summary, we propose an evolutionary attack that supersedes SOTA attacks within global, local, and targeted regimes. Unlike gradient-based methods, \method can use additional computational resources to perform even better. And since it does not require differentiability, it can extend to other objectives including two new targets: conformal guarantee, and certified ratio. Our results clearly show that the space of evolutionary (and more broadly search-based) attacks has a lot of untapped potential.

\section{Background and Related Work}
\label{sec:backgournd}
\parbold{Problem setup} We focus on attacking the semi-supervised node classification task via perturbing a small number of edges. We are given a graph $\gG = (\mX, \mA, \vy)$ where $\mX$ is the features matrix assigning a feature vector $\vx_i$ to each node $v_i$ in the graph, $\mA$ is the adjacency matrix (often sparse) representing the set of edges $\gE$, and $\vy$ is the partially observable vector of labels. Nodes are partitioned into labeled and unlabeled sets $\gV = \gV_l \cup \gV_u$. The GNN is trained on a clean initial subgraph $\Gtr$ that includes the labeled nodes.
Following \citet{lrbcd} we avoid the transductive setup ($\Gtr = \gG$) since perfect robustness can be achieved there by only memorizing the clean graph during training. They show that adversarial and self training also show a false sense of robustness in that setup for the same reason. Instead, we focus on inductive learning where a model $f$ is trained on an induced subgraph $\Gtr \subseteq \gG$, validated on $\Gval \subseteq \gG$ and tested on 
$\Gtest$ where $\Gtr \subset \Gval  \subset \Gtest = \gG$.

\parbold{Threat model}
Our goal is to find a perturbation matrix $\mP \in \{0, 1\}^{n\times n}$ that flips entities of the adjacency matrix $\tilde\mA = \mA \oplus \mP$, where $n = |\gV|$, and $\oplus$ is the element-wise XOR operator. We optimize over $\mP$ to decrease the accuracy. For a given function $f$ as the GNN model, and any generic loss function $\gL$, the objective is
\begin{equation}
\label{eq:attack}
      \mP = \underset{{\mP}}{\argmax}   \quad  {\mathcal{L}}(f(\gG(\mX, \mA\oplus \mP))_{\mathrm{att}}, \vy_{\mathrm{att}}) \qquad
        s.t.  \quad \1_{N}\mP \1_N^\top \leq \epsilon \cdot |\gE[\Vattack:\gV]| 
\end{equation}
Here $f(\cdot)_{\mathrm{att}}$ is the vector of predictions for the subset of nodes  $\Vattack$ that are under attack. In ``targeted'' attacks $\Vattack$ is a singleton.
To keep the perturbations imperceptible, we assume that the adversary can only perturb up to $\delta:= \epsilon \cdot |\gE[\Vattack:\gV]|$ edges where $\gE[\gA:\gB]$ is the subset of edges between nodes in $\gA$ and $\gB$.
 \autoref{eq:attack} can include more constraints like the local constraint from \citet{gosch2023adversarial} restricting perturbations not to increase node degrees by more than a fraction (e.g $e_\mathrm{loc} = 0.5$) of their original value. \looseness=-1

\parbold{Related Work}
We study evasion attacks with both global and targeted objectives, where perturbations are introduced only at test time. The goal is either to reduce the model’s overall accuracy or to induce the misclassification of a specific node. Among gradient-based methods, PRBCD \citep{prbcd} and LRBCD \citep{lrbcd} represent the current state of the art. Both attacks compute gradients of the $\tanh$-margin loss with respect to the adjacency matrix, employing block-coordinate descent. Perturbation edges are then sampled based on these gradients. To handle local degree constraints, LRBCD incorporates a local projection step, greedily selecting edges (in descending order of gradient score) while ensuring that the constraints are not violated.

Beyond gradient-based methods, alternative approaches have also been explored. For example, \citet{dai2018adversarial} proposed a simple evolutionary attack as a baseline for their reinforcement learning-based method. However, these early strategies have since been surpassed by gradient-based techniques. Building on this progress, we redesign key components of the search process, achieving significant improvements over prior evolutionary attacks. Moreover, our method, \method, scales effectively to large graphs and naturally extends to novel attack objectives. Other heuristic-based attacks, relying on node degree, centrality, or related metrics \citep{10448076, 10.1145/3611307, 10.1145/3589291}, have also been proposed, but they similarly fail to outperform the current state-of-the-art methods. A more detailed discussion of related work can be found in \autoref{sec:related-works-extended}.

\section{\method: Evolutionary Attack}
\label{sec:method}

\method use genetic algorithm \citep{holland1984genetic} as a meta-heuristic search to directly optimize \autoref{eq:attack}. We start with an initial set of possible (candidate) perturbations, called a ``population'', and iteratively improve it by removing non-elite candidates and fusing elites to form new perturbations. The candidates in a population are sorted based on their ``fitness'' which for global and local attacks is directly set to the accuracy. The next population is drafted by keeping the best individuals and producing new ones from the top candidates through a process called ``cross-over''. To introduce exploration, we add random changes to each candidate with a ``mutation'' operator. \looseness=-1

\parbold{Components of \method} We define a genetic solver with of four main components. \begin{enumerate*}[label=(\roman*)]
    \item Population: a set of feasible answers to the problem that gradually improve over iterations. Here, each candidate is a perturbation to the original graph, a vector of indices at which an edge will flip; formally $\vs_{i} \in [\frac{n}{2}(n-1)]^\delta$.  Indices are calculated via a mapping $\Pi:  [n]^2 \mapsto [\frac{n}{2}(n-1)]$ that is an enumeration on the upper triangle of the $n\times n$ adjacency matrix (see \autoref{sec:technical}).
    The corresponding perturbation matrix $\mP_{i}$ is simply defined as $\mP_{i}[p_t, q_t] = \mP_{i}[q_t, p_t]= 1$ where $(p_t, q_t) = \Pi^{-1}(\vs_{i}[t])$ for every index $t$. The initial population is selected randomly.
    \item Fitness: is a notion of how close to optimal each candidate is. For any loss function $\gL$ we define the fitness function $\mathrm{fit}:[\frac{n}{2}(n-1)]^\delta \mapsto \sR$, as $\mathrm{fit}(\vs_i) = \gL(\mX, \mA \oplus \mP_i, \vy)$. 
    Regardless of differentiability, as long as the loss function has enough sensitivity to contrast between various individuals, we use it directly to compute the fitness (special case in \autoref{sec:local-and-targetted}). 
    \item Crossover: is an operation to generate new population candidate via combining two existing ones. The (single joint) crossover operation at joint $j$ defines a new candidate vector $\vs_\mathrm{new} = \mathrm{cross}_j(\vs_1, \vs_2) := \vs_1[:j]\bullet\vs_2[j+1:]$ where $\bullet$ is the concatenation of two vectors. Crossover operation with $k_\mathrm{cross} > 1$ joints is defined recursively in the order of joints. The number of crossovers is a hyperparameter (see \autoref{sec:dataset-models-hyperparameters}), and their locations is chosen randomly in the range of canididates' length. 
    The candidates for cross-over are chosen through a ``tournament''. In each tournament, $n_\mathrm{tour}$ random candidates are compared, and the parent candidates are selected based on their fitness. This process repeats for $t$ generations.
    \item Mutation: introduces further exploration to the new population. The function $\mathrm{mutate}: [\frac{n}{2}(n-1)]^\delta \mapsto [\frac{n}{2}(n-1)]^\delta$ is a random mapping of a candidate to another. A simple example of mutation is to changes each index with some mutation probability $p$ to any random index in the range. We propose better mutation operators later.
\end{enumerate*}

\parbold{Sparse encoding of the attack} A simple way to represent a perturbation is a boolean vector of size $N^2$ encoding which edges are flipped. It costs $\gO(|\gS|N^2)$ space from memory where $\gS$ is the population.
This representation is not aware of the sparsity in $\mA$. Instead we represent each candidate as a list of indices to be toggled in the adjacency matrix --- we store sparse representation of $\mP$.
With this we account for the sparsity and reduce the complexity to $\gO(|\gS| \cdot \delta)$ where $\delta = \mathfloor{\epsilon \cdot |\gE[\Vattack:\gV]|}$ -- candidates in the population $\vz \in \gS$ are vectors of $\delta$ dimensions with each entity as an index in adjacency matrix $\vz[i] \in \{1, \cdots, n(n-1)/2\}$ with $n = |\gV|$. 
Our mapping $\Pi$ (as discussed in \autoref{sec:technical}) is a diagonal enumeration of an upper triangular $n\times n$ matrix. For simplicity, we let the perturbation vector to contain repeated elements. During the evaluation of the vector, we transform it to a perturbation matrix $\mP_\vz$, with which we compute $\tilde\mA = \mA \oplus \mP_\vz$. We compute all steps with sparse representation, where each candidate takes $\gO(\delta)$ space. Moreover, with this encoding, we directly enforce the global budget since the size of each individual in the population at most the number of allowed perturbations by design. \looseness=-1

\begin{figure}[t]
  \begin{center}
  \input{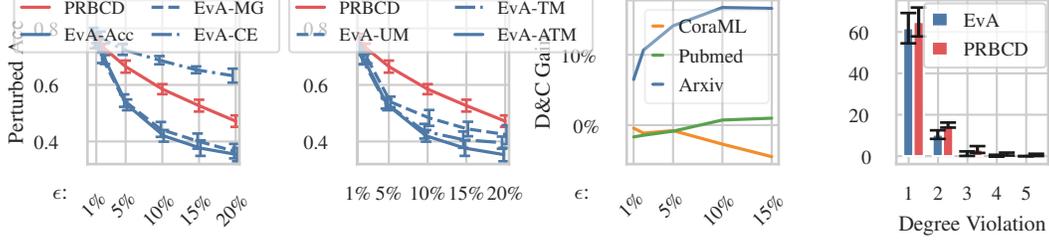}
  \end{center}
  \caption{[Left] \method's performance with different objective functions and [Middle left] mutation strategies. [Middle right] Effect of D\&C on the performance of \method for different datasets. [Right] The number of violations from local constraints by EvA, and PRBCD ($\epsilon_\mathrm{loc} = 0.5$).}
  \label{fig: objective_mut}
\end{figure}

\parbold{Accuracy vs alternative surrogate losses}
To understand the effect of the loss on the attack, we conducted an ablation study to compare accuracy and common surrogate objectives (cross-entropy, and margin-based loss) as the fitness function in \method. As in \autoref{fig: objective_mut} (left), cross-entropy does not use the attack budget effectively, while margin-based loss shows to be well-correlated. Intuitively, since the goal is to misclassify as many nodes as possible, the aggregated cross-entropy loss can waste perturbation budget by overly focusing on already misclassified nodes, rather than maximizing new misclassifications. This effect was studied in depth by \citet{geisler2023robustnessgraphneuralnetworks}, which motivated them to introduce the {tanh-margin} loss which mitigates this issue. Still among all fitness functions, accuracy itself performs better. Since PRBCD uses the tanh-margin loss, the large gap between \method, and PRBCD suggests that the quality of loss is not the only reason behind \method's effectiveness. We hypothesise that \method, leveraging the exploratory capabilities of GA, can explore the search space more effectively and avoid local optima, while PRBCD gets stuck.

\parbold{Drawbacks} The mentioned setup is the baseline variant of \method. Combined with cross-entropy as the fitness it is similar to a parallel and efficient implementation for \citet{dai2018adversarial} which is by far less effective (see \autoref{fig: motivation_figure}). 
While the baseline (with accuracy) already outperforms SOTA \autoref{fig: objective_mut}, we enhance the search by introducing a better initial population and a mutation function that discards edges outside of the target's receptive field.

\parbold{Enhancing the search}
To enhance \method, the key insight is that by restricting the search space to the receptive field of $\Vattack$ (instead of the entire $\frac{n}{2}(n-1)$ edges), we eliminate less effective (or ineffective) perturbations from the search space. Perturbations that have both endpoints in the training subgraph can be easily reverted by memorization. Additionally, flipping edges outside of the receptive field of $\Vattack$ is a waste of budget since they do not affect the prediction of $\Vattack$.
Similarly, we restrict the initial population to have at least one endpoint in $\Vattack$. This is easily done by randomly sampling both endpoints, one inside $\Vattack$ and one in $\gV$, then mapping the edges back to the indices via $\Pi$.
For larger graphs, as the search space increases quadratically to the number of nodes, we can apply a divide and conquer strategy by splitting $\Vattack$ and running \method on each chuck.

\parbold{Targeted and adaptive mutation} 
Mutation is applied by selecting a set of perturbation indices (uniformly at random with probability $p$) from the population and changing them to another index. A naive implementation (the \underline{u}niform \underline{m}utation (UM)), adds random indices from anywhere in the entire graph. 
Similar to the initialization, we define the ``\underline{t}argeted \underline{m}utation (TM)'' by restricting the new mutated edge to have at least one end-point in $\Vattack$. Furthermore, when the attack succeeds in altering a node's label, perturbing its connections does not increase the performance anymore. Hence, we exclude  the already flipped nodes from the endpoint that was restricted to $\Vattack$. Importantly, we still let those nodes to connect with other nodes in $\Vattack$ as they can contribute to the misclassification risk of other nodes. We refer to this approach as ``\underline{a}daptive \underline{t}argeted \underline{m}utation'' (ATM).  Remarkably, as shown in \autoref{fig: objective_mut} (right), these modifications improve the effeciveness of \method by a noticable margin. \looseness=-1

\parbold{Stacking perturbations} \method requires a forward pass per each candidate (each candidate of population). 
While maintaining the sparse representation of the graphs during all steps, we can use the remaining memory to combine $k$ candidates in form of a large graph of $k$ parts and evaluate all $k$ in a single forward pass. 
In practice, we can easily fit the entire population in one forward pass per iteration as one large graph.

\parbold{Effect of scaling} 
The population size has a considerable impact on the performance of \method by introducing diversity among the solutions, thus increasing exploration.
To observe this effect, we do an ablation study on the population size and the number of iterations. For a fair comparison, we scale PRBCD separately by increasing the number of steps and the size of the block coordinate subspace. We exponentially increased the block size, starting from 500 up to 4 million.
\begin{wrapfigure}{r}{0.325\textwidth}
    \centering
        \vspace{-2.4em}
    \input{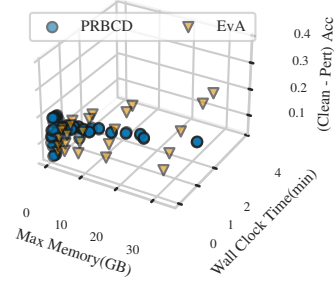}
         \caption{Wall-clock time vs. memory tradeoff for \pubmed.}
        \label{fig: scaling-time}
\end{wrapfigure}
As shown in \autoref{fig: scaling} (left), increasing the population size, and then increasing the number of iterations improves \method's in that order. 
In contrast, PRBCD does not achieve noticable improvement by increasing the block size or the number of training steps (\autoref{fig: scaling} (mid)).
This means that \method leverages additional computational resources (either time or memory) while PRBCD does not show a considerable use of it. As a supplement to \autoref{fig: objective_mut}, in \autoref{fig: scaling} (right) we show that using a better mutation function (here adaptive targeted mutation) consistently enhances performance across all population sizes and outperform the uniform approach. We compare \method and SOTA for wall-clock time and memory in \autoref{fig: scaling-time} showing that due to the exploratory nature of \method, it saturates later. In general, we see that \method is more often Pareto optimal, and a broader range of time-memory-performance trade-offs.

\parbold{Divide and Conquer}
PRBCD uses the coordinate gradient descent to scale efficiently for larger graphs. Similarly we introduce a divide-and-conquer (D\&C) approach to \method. Here instead of attacking the entire $\Vattack$ at once, we divide it into smaller subsets and sequentially attack each subset with a budget relative to the portion of the edges connected to it.
After attacking a subset, we treat the modified graph as a starting point for the next one. The result for the final subset includes perturbations in all previous steps. At the end we re-evaluate the final graph with all perturbations combined. 
Our divide and conquer approach relies on a relaxation. For a budget of $\delta=\delta_1 + \delta_2$ over a set $\Vattack = \gV_1 \cup \gV_2$, the standard attack searches in the space of $\binom{n}{2}^\delta$ possible perturbations aiming to decrease the accuracy over $\Vattack$. However, with the divide and conquer approach, the attack searches among $\binom{n}{2}^{\delta_1} + \binom{n}{2}^{\delta_2}$ possible perturbations each aiming to attack $\gV_1$, and $\gV_2$ separately - first searching for optimal attack with a budget $\delta_1$ on $\gV_1$ and then with $\delta_2$ on $\gV_2$ given the attack applied on $\gV_1$. Therefore, D\&C explores an exponentially smaller subset of the search space. This calculation is for the uniform mutation, employing targeted mutation narrows the search space explored by the algorithm further. Since it also reduce the choices from $\binom{n}{2}^\delta$ to $(\frac{|\Vattack|(2n-|\Vattack|-1)}{2})^ \delta$.
In practice we divide $\Vattack$ to $k_\mathrm{dc}$ subsets (see hyper-paramters in \autoref{attack-hyperparameters}). 
Applying the D\&C approach poses a trade-off: in smaller spaces \method finds better solutions, while the relaxation in D\&C can lead to solutions further from optimum. As shown in \autoref{fig: objective_mut} (right) when the size of the graph and the budget $\delta$ grow, adding D\&C to \method helps substantially. As in \autoref{fig: objective_mut} (right) it improves the result for large \arxiv by at least $\sim 8\%$ while the same approach is ineffective for smaller \coraml dataset. We further show that on large graphs D\&C similarly helps PRBCD. Indeed, applying D\&C for PRBCD helps to increase the performance while maintaining the same block-size (not increasing the required memory; see \autoref{sec:d-and-c}). A comparable block for 1-step PRBCD exceeds the memory limit. 
It is noteworthy that the randomized block-coordinate computation of gradients is also a relaxation.

\begin{figure*}[t]
    \begin{minipage}[t]{0.56\textwidth}
    \centering
    \resizebox{\textwidth}{!}{
    \input{./figures/fig_scaling.pgf}
    }
       \caption{(From left to right) scaling performance \\ of \method and PRBCD over various memory and  \\iteration budgets, the effect of mutation on different \\ resources with $\epsilon=0.1\%$ on \pubmed.\looseness=-1}
       \label{fig: scaling}
    \end{minipage}
    \hfill
    \begin{minipage}[t]{0.43\textwidth}
    \centering
    \resizebox{\textwidth}{!}{
    \input{figures/histplot_targeted_attack.pgf}
    }
    \caption{Number of perturbations (different colors) used by EvA [left] and PRBCD [right] to attack nodes of specific degrees. Black (NA) shows failed attacks.}
   \label{fig:targetted-attack}

    \end{minipage}
\end{figure*}

\section{Local and Targetted Attacks \& Attacking Other Objectives}
\label{sec:local-and-targetted}

\parbold{Local attacks} \citet{lrbcd} extend PRBCD to support additional ``local'' constraints where a perturbation is not allowed to increase the degree of a node more than a fraction of its original value.
We need local constraints to enforce imperceptibility of the attack. For example, a perturbation might increase the degree of a node more than twice its original value while staying within the global budget. Therefore, even within the global budget the structure of the graph (and therefore graph’s structural semantics) can change drastically.
They introduce the LRBCD attack which adds a local projection to PRBCD. In short, they sort edges in a decreasing order of probability (gradients), and iteratively add perturbations while the local constraint for both end-points of the modified edge is not violated. This continues until the global budget is exhausted. Even without enforcing this restriction, as shown in \autoref{fig: objective_mut} (right) \method introduces less degree violations compared to PRBCD, meaning that the perturbations added by \method are more spread-out in the graph.
We apply a local projection to \method similar to LRBCD. Here, instead of using gradients for ranking, we use the frequency of the edge within the current population. We define $s(e) = \sum_{\vs\in \gS}\sI[e\in\vs] / |\gS| + u$ as the frequency score where $u$ is a small uniform random value in $[0, 0.05]$. Here $u$ is added to break ties and introduce additional randomness, and $\gS$ is the population at the current iteration. Our insight is that if an edge appears frequently within the population, it is likely to be useful for an attack, increasing the chance of candidates containing it to be selected as elite.  After our local projection all constraints are guaranteed to be satisfied.
For more diversity at initial iterations, we apply a random projection removing edges with a probability proportional to total degree violations on both sides. We apply this random removal for $t_\mathrm{warm}$ iterations (a hyper-parameter discussed in \autoref{attack-hyperparameters}).\looseness=-1

\parbold{Node-targeted attacks} 
Here the objective is to misclassify a specific node with as minimal change to the structure as possible. Using \method with the global setup does not work in this case. On a single node the accuracy has only two values $0$ or $1$ -- small changes in the solution do not result in (even minor) changes in the fitness score.
With the 0-1 accuracy objective, random search and GA are practically equivalent as there is no indication of what combination of edges are closer to breaking the prediction of a particular node -- all non-successful combinations are equally evaluated with 1. 
Instead we use the proxy $\tanh$-margin loss as the fitness function. This loss function changes as we perturb the receptive field of the targeted node. Note, for general (non-targeted) attacks the $\tanh$-margin loss improves performance over the cross-entropy loss, however, using accuracy (for larger $|\Vattack|$) as fitness is slightly better as shown on \autoref{fig: objective_mut} (left).
\autoref{fig:targetted-attack} compares \method, and the state-of-the art attack PRBCD on targeted attacks. \looseness=-1

\parbold{Other objectives}
For non-differentiable objectives (e.g. accuracy), gradient-based attacks need a differentiable surrogate approximating it. As discussed in \autoref{sec:method}, for accuracy (common setup) several works proposed various surrogates. This is similarly challenging to propose gradient-based attacks for novel objectives that are complicated and include several non-differentiable components (e.g., quantile computation or majority voting from Monte Carlo samples). Since our method nullifies the need for information from gradients, we can easily optimize for novel complex objectives as long as they are sensitive to small changes in the search space. We define three new attacks on graphs: reducing the certified ratio of a smoothing-based model, decreasing the coverage, and the increasing set size of conformal sets. A detailed explanation of randomized smoothing-based certificates and conformal prediction, which underpin the certified ratio objective and conformal prediction, is provided in \autoref{app:smoothing_certificates} and \autoref{app: conformal prediction} respectively.

\parbold{Attacking smoothing-based certificate} 
Assuming the certified ratio is a notion of a trustworthy prediction, one possible adversarial objective is to reduce the number of nodes that are certified (a.k.a. certified ratio) while maintaining the same clean accuracy. While the operations includes non-differentiable steps we can directly set the certified ratio (fraction of nodes that are certified within a determined threat model) as the objective of \method. Whether a node is certified reduces to whether the smooth classifier returns a probability above $\overline{p}$ where $\min_{\tilde\vx \in \gB(\vx)} g(\vx) \ge 0.5$ constrained to $g(\vx) = \overline{p}$. Many smoothing-based certificates are computed at canonical points (they are only a function of probability not the input) and they are non-decreasing to $\overline{p}$. Hence, we find $\overline{p}$ via binary search. Thus, our objective is to decrease the number of vertices with smooth probability above $\overline{p}$. A naive implementation of \method for this objective is to compute the certified ratio given new MC samples for each candidate. This increases the runtime of our algorithm by a factor of $n_\mathrm{MC}$, as each perturbation requires $n_\mathrm{MC}$ forward passes.
Inside the attack, statistical rigor is not crucial. Therefore, we employ an efficient sampling strategy where we start with initial samples from clean $\mA$, and for each perturbation, we only resample for the edges in $\tilde\mA\oplus\mA$. We use the stacked inference technique (see \autoref{sec:method}) on MC samples which ultimately reduces the computation to one inference per each perturbation $\tilde\mA$. 

\parbold{Attacking conformal prediction} A common threat model for CP is to decrease the empirical coverage (far from the guarantee) by perturbing the test input. We propose a similar attack where the adversary changes the edge structure of the graph in order to decrease the coverage. 
This process is again not directly differentiable (for steps like computing the quantile and comparison of values) which is not a problem for \method.
In our experimental setup, the defender calibrates on a random subset of $\gV_u$ (besides the test, this is the only set with labels unseen by the model). Assuming that the unlabeled and test nodes are originally exchangeable (node-exchangeability), the conformal guarantee is valid in the inductive setup upon recalibration on the clean graph. By perturbing the edge structure we can easily break this guarantee. Therefore our objective is to change the edge structure such that the coverage is minimized. Intuitively, this requires maximizing the distribution shift between the test and calibration scores. 
As we know that the calibration set is an exchangeable (random) subset of $\gV_u$, we set the entire $\gV_u$ as the calibration set during the attack. Due to exchangeability we expect the similar effect from our perturbation for any random subset as well \citep{berti1997glivenko}. Finally, the objective is to decrease the coverage over $\Vattack$ given $\gV_u$ as the calibration set.
To the best of our knowledge, so far this is the only adversarial attack on the graph structure to break conformal inductive GNNs. Similarly, by changing the objective to the negative average set size, we can attack the usability of prediction sets (see \autoref{fig: conformal_certificate_attack}). \looseness=-1

\section{Empirical Results}
\label{sec:emp_res}

With our empirical evaluations we show that current gradient-based attacks are still very far from optimal since \method outperforms them by a notable margin. \method inherently results in attacks with smaller local change in each nodes' degree (even without posing local constraints)\autoref{fig: objective_mut}[right]. And further we can apply \method to attacks with local constraints as well. 
    With divide and conquer, \method is able to scale to larger graphs (e.g. \arxiv) and outperform SOTA for those graphs as well.
    With the black-box nature of the attack we easily extend the score of \method to novel objectives introducing the first attack to reduce the certified ratio or break conformal sets on graphs. 

\begin{figure}
    \centering
    \input{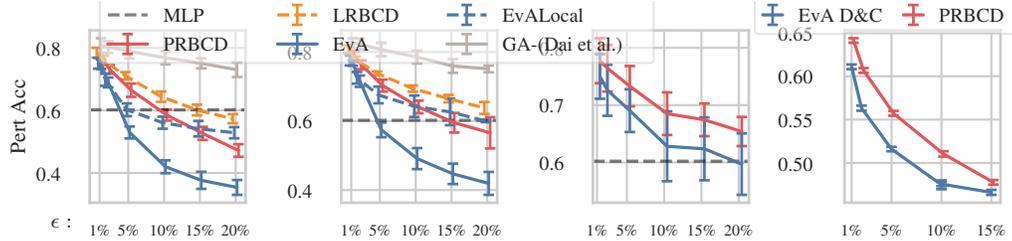}
    \caption{(Left to right) Performance on \coraml on Vanila GCN, adversarially trained GCN using PRBCD, Soft-Median-GDC model. The right-most figure is GCN on \arxiv.}
    \label{fig:all-vanilla-gcn}
\end{figure}

\parbold{Experimental setup}
We evaluate \method on common graph datasets: \coraml \citep{cora}, \citeseer \citep{citeseer}, and \pubmed \citep{pubmed}. \citet{pitfalls} show that GNN evaluation is sensitive to the initial train/val/test split. Therefore, we averaged our results for each dataset/model over five different data splits. 
In contrast with common GNN attacks, \citet{lrbcd} show that transductive setup carries a false sense of robustness. In other words, trivially one can gain perfect robustness just by memorizing the clean graph which is available before the attack; models with robust and self-training also show to exploit this flaw. Following them, we report our results in an inductive setting. We divide graph nodes into four subsets: training, validation, and testing, each with 10\% of the nodes and we leave the remaining 60\% as unlabeled data. 
Following \citet{lingam2023rethinking}, we sample the train, validation and test nodes in exchangeably since it provides a more realistic setup compared to commonly used methods, such as sampling for training and validation with the same count for each class (i.e. stratified sampling).
For completeness, in \autoref{sec:experiments-appendix} we compare attacks in the transductive setup and various sampling approaches. In all cases again \method shows a more effective attack. Further information about the model and hyperparameters are in \autoref{sec:dataset-models-hyperparameters}.

\parbold{Attacking vanilla and robust models} As shown in \autoref{fig:all-vanilla-gcn} and extensivelt in \autoref{sec:experiments-appendix}, \method outperforms the SOTA attack PRBCD by a significant margin consistently across various datasets and models (vanilla and robust). 
Interestingly, we show that on many vanilla and robust models, for a very small budget $\epsilon \sim 0.05$, \method drops the accuracy below the level of the MLP model. This is a condition where the model leveraging the structure works worse than a model that completely ignores edges. The SoftMedian model seems to show an inherent robustness to both \method and PRBCD. Therefore, to break the model below the accuracy of MLP, we require $\ge 0.2$ perturbation budget. Even in the SoftMedian model, our attack is significantly more effective in comparison to PRBCD.
As expected, models trained with \method were shown to be more robust, and other attacks were less effective to them. However, this additional robustness is not significant. \autoref{tab:coraml_inductive} (\autoref{sec:experiments-appendix}) compares attacks in models with different adversarial training. Further, we also study the perturbed edges in \autoref{app:labeldiversity}.

\parbold{Scaling to larger graphs}
In \autoref{fig:all-vanilla-gcn}, we also show that the \method applied with our divide and conquer approach outperforms PRBCD for \arxiv dataset. Interestingly similar divide and conquer approach can significantly improve PRBCD as well; while still \method is more effective. In \autoref{sec:d-and-c}, we compared PRBCD with block size 3M, 10M, alongside PRBCD and \method with divide and conquer in a fair comparison. 
Notably PRBCD with the highest block size fitting in one GPU is still significantly less effective compared to any of the attacks combined with D\&C.

\begin{minipage}{0.3\textwidth}
\parbold{Additional datasets}
 To show that \method generalizes beyond citation graphs we compare it with PRBCD on the \textsc{Amazon-Photo} and \textsc{Amazon-Computers} graph \citep{pitfalls} in \autoref{tab:non-citationgraph}. \method is still better.
\end{minipage}
\hfill
\begin{minipage}{0.66\textwidth}
\setlength{\tabcolsep}{2pt}
\centering
\captionof{table}{{Performance on non-citation graphs.}}
\resizebox{\textwidth}{!}{
\label{tab:non-citationgraph}
\begin{tabular}{llcccc}
\toprule
Dataset & Attack & $\epsilon=0.01$ & $\epsilon=0.02$ & $\epsilon=0.05$ & $\epsilon=0.10$ \\
\midrule
\multirow{2}{*}{\texttt{photo}} 
& PRBCD & 84.28 {\scriptsize$\pm$ 0.99} & 81.12 {\scriptsize$\pm$ 1.45} & 75.86 {\scriptsize$\pm$ 1.81} & 71.58 {\scriptsize$\pm$ 1.80} \\
& \method & \textbf{80.43 {\scriptsize$\pm$ 1.47}} & \textbf{77.99 {\scriptsize$\pm$ 2.04}} & \textbf{72.73 {\scriptsize$\pm$ 1.90}} & \textbf{67.01 {\scriptsize$\pm$ 1.76}} \\
\midrule
\multirow{2}{*}{\texttt{computers}} 
& PRBCD & 77.28 {\scriptsize$\pm$ 0.68} & 73.56 {\scriptsize$\pm$ 0.55} & 66.92 {\scriptsize$\pm$ 0.63} & 61.99 {\scriptsize$\pm$ 0.59} \\
& \method &\textbf{ 72.94 {\scriptsize$\pm$ 1.26}} &\textbf{ 70.10 {\scriptsize$\pm$ 1.85}} &\textbf{ 65.70	 {\scriptsize$\pm$ 2.89} }&\textbf{ 60.13 {\scriptsize$\pm$ 3.72}} \\
\bottomrule
\end{tabular}
}
\end{minipage}

\parbold{Local attacks} Similarly, as shown in \autoref{fig:all-vanilla-gcn}, and \autoref{sec:experiments-appendix}, \method is consistently better than LRBCD. In \autoref{sec:local-and-targetted} discussed that we apply local projection as a mutation function. Interestingly as in \autoref{fig:divide-conqure} (right) even without local projection, \method results in less violations of the local constraint.

\parbold{Targeted attack} We perform targetted attacks on each node separately, with varying budgets from one to a maximum of 10 edges, until the prediction changes. We discussed in \autoref{sec:local-and-targetted}, that here we used $\tanh$-Margin proxy loss since accuracy on one node is not sensitive to small changes. \autoref{fig:targetted-attack} compares \method and PRBCD in tagetted attack. Our results show that PRBCD performs better with a budget of one, but is outperformed by \method for budgets of two and higher. For instance, on the \coraml dataset PRBCD fails to modify 16 nodes with a maximum of 10 changes (NA, black), whereas this number is reduced to only 2 nodes for \method. This result is expected due to the combinatorial nature of the problem: for budgets up to two, a greedy approach can find the optimal solution, but as the budget increases beyond three, the problem becomes significantly more complex. This is also in line with our first motivation that the gradient ignores the interaction effect of flipping multiple edge simultaneously. \autoref{fig: motivation_figure} [middle] is an instance when the gradient direction individually has the same direction, but the loss when flipping both is in the opposite direction. This effect can become even more problematic when one flips more edges.

\begin{figure}
    \centering
    \input{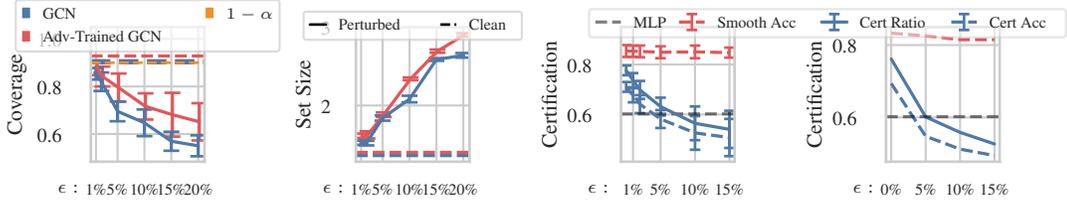}
    \caption{(From left to right) conformal coverage, and conformal set size on vanilla and adversarially trained GCN. 
    The certificate attack for certified ratio on $
    \mA$ and $\mX$ evaluated on GPRGNN adversarially trained using PRBCD. All plots are for \coraml.}
    \label{fig: conformal_certificate_attack}
\end{figure}

\parbold{Attacking novel objectives}
In \autoref{fig: conformal_certificate_attack} (mid-right and right) we performance of \method with the objective to reduce the certified ratio. The plots are for certificate on $\mA$ (mid-right) with $(p_+=0.001, p_-=0.4)$, and $\mX$ (right) with $(p_+=0.01, p_-=0.6)$ with sparse smoothing \citep{bojchevski2020efficient}. Here $p_+$, and $p-$ are Bernoulli parameters of flipping a zero or one. In both plots we report the result for $\gB_{0, 3}$ which means 0 additions and 3 deletions.
While we aim to decrease the certified ratio, a direct outcome is that the certified accuracy drops. For a 5\% budget, the certified accuracy drops below MLP. 
MLP is a baseline model with full robustness to edge perturbations (since it discards the adjacency information completely). While reducing the certified ratio, interestingly the smooth model's accuracy remains the same. Hence, evaluating the model on a holdout labeled set does not reveal that the input graph is attacked. 
We report the first structure attack to inductive conformal GNN. As shown in \autoref{fig: conformal_certificate_attack} (right) the coverage drops quickly as we increase the perturbation budget. As expected, in an adversarially trained model, we observe a slower decrease in the empirical coverage. Alternatively in \autoref{fig: conformal_certificate_attack} (middle) we increase the average set size since showing that that both vanilla and robust models are vulnerable to this attack.

\parbold{Ablation study on the effect of our different GA extensions}
To emphasize the effects of our simple yet effective enhancements, we provide the following ablation studies.
In \autoref{tab:abl_comp1}, we show the effect of each enhancement individually and then together (\method) on the \texttt{\coraml} dataset. Furthermore, in \autoref{tab:abl_comp2}, we report the effect of our sparse encoding (SE) and D\&C on the larger \arxiv dataset. As shown, all of our simple enhancements provide a significant effect (both separately and jointly).

\begin{minipage}{0.5\textwidth}
\setlength{\tabcolsep}{4pt}
\captionof{table}{The effect of our adaptive targeted mutation (ATM) and the fitness function on \coraml.}
\centering
\resizebox{\textwidth}{!}{
\begin{tabular}{lccccc}
\toprule
\textbf{$\epsilon$} & \textbf{0.01} & \textbf{0.02} & \textbf{0.05} & \textbf{0.10} & \textbf{0.15} \\
\midrule
$(*)$ Dai et al. & 80.71 & 80.28 & 78.86 & 76.86 & 75.08 \\
$(*)$ + ATM & 78.50 & 76.65 & 72.52 & 68.75 & 65.33 \\
$(*)$ + $\gL_\mathrm{acc}$  & 75.08 & 69.39 & 54.02 & 48.32 & 44.41 \\
EvA (+ both) & 74.80 & 68.96 & 52.95 & 41.99 & 37.65 \\
\bottomrule
\end{tabular}
}

\label{tab:abl_comp1}
\end{minipage}
\hfill
\begin{minipage}{0.45\textwidth}
\setlength{\tabcolsep}{4pt}
\captionof{table}{Effect of our sparse encoding and D\&C on the large \arxiv dataset.}
\resizebox{\textwidth
}{!}{
\centering
\begin{tabular}{lcccc}
\toprule
$\epsilon$ & \textbf{0.01} & \textbf{0.02} & \textbf{0.05} & \textbf{0.10} \\
\midrule
Dai et al. & OOM & OOM & OOM & OOM \\
Dai et al. + SE & 69.79 & 69.56 & 68.81 & 67.77 \\
EvA & 66.86 & 66.80 & 65.18 & 63.51 \\
EvA + D\&C & 61.08 & 56.31 & 51.60 & 47.56 \\
\bottomrule
\end{tabular}
\label{tab:abl_comp2}
}
    
\end{minipage}

\section{Conclusion}
In contrast to gradient-based adversarial attacks on graph structure, we developed a new attack (\method) based on a heuristic genetic algorithm. 
By eliminating differentiation, we can directly optimize for the objective of the adversary (e.g. the model's accuracy). 
This black-box nature enables us to define complex adversarial goals, including attacks on robustness certificates and conformal prediction. Our novel attacks decrease the certified ratio, and conformal coverage, and increase the conformal set size. 
We propose an encoding that reduces the memory complexity of the attack to the same order as the perturbation budget which allows us to adapt to various computational constraints. 
For larger graphs by introducing a divide and conquer approach we increased the performance of both PRBCD and \method (while still \method outperforms). We also show that due to the open-ended characteristic of the search, for more computational resources (time and memory) we can always improve our results.
Given the drastic decrease in the model's accuracy by applying \method, we highlight that even SOTA gradient-based attacks are far from optimal. Our main message is that search-based attacks are underexplored yet powerful as shown by our results. Our code is submitted in the supplementary matrial. \looseness=-1

\parbold{Limitations} We use an off-the-shelf genetic algorithm. Surely, there is room for designing search algorithms specific to the domain of the problem beyond our extensions, or even hybrids of gradient and evolutionary search.  
\method uses many forward passes through the model which can be unrealistic in some attack scenarios. We leave the design of a further query-efficient variant for the future.

\bibliographystyle{plainnat}

\bibliography{refrence}

\newpage
\appendix

\section{Supplementary to Related Work}

\label{sec:related-works-extended}
We focus on evasion attacks where perturbations are made after the model's training. Based on the domain, (evasion) attacks can be further be categorized to global (aiming to flip the prediction of a subset of nodes) and targeted attacks (aiming at a single node). Our attack applies on edge-structure similar to \citet{xu2019topology, zugner2018adversarial, geisler2023robustnessgraphneuralnetworks, prbcd, lrbcd}. Orthogonal to this scope, various other grpah attacks are proposed in the literature including node-injection attacks \citep{ju2023let}, poisoning \citep{zugner2020adversarial, lingam2023rethinking, zugner2018adversarial}, and attacking attributes \citep{zugner2018adversarial}. 
Inspired by techniques used on continuous data, \citet{xu2019topology, zugner2018adversarial, geisler2023robustnessgraphneuralnetworks}  utilize gradients to approximate perturbations on inherently discrete edges. As the adjacency matrix can grow significantly larger than images, applying a PGD-like attack becomes challenging for larger graphs. To remedy that \citet{prbcd} proposes a block-coordinate computation of the derivatives, and \citet{lrbcd} applies a greedy projection to apply local constraints.

Orthogonally, \cite{rl_attack} use reinforcement learning to refine their attack and disrupt the learning process of GNNs \cite{9878092}. They also introduce a genetic algorithm attack as a baseline; however, they did not design the components of GA carefully. In \autoref{sec:method} we design GA components (mutation, local projection, etc) which outperform recent gradient based attacks. Recently new attacks relying on heuristics such as node degree, centrality, etc have been proposed (e.g. \citet{10448076, 10.1145/3611307, 10.1145/3589291}), however they don't outperform the SOTA.

\parbold{Gradient-based attacks} A common class of attacks compute the gradient of the objective w.r.t. $\mA$. This requires a relaxation on the domain of $\mA$ from $\{0, 1\}^{n\times n}$ to $[0, 1]^{n\times n}$. For non-differentiable objectives like accuracy differentiable surrogates like the categorical cross entropy or $\tanh$-margin \citep{geisler2023robustnessgraphneuralnetworks} are used instead. The algorithm is to iteratively compute the gradients and update the perturbation matrix. Finally, based on the continuous perturbation matrix edges are either sampled or rounded to the binary domain. 

\parbold{Black-Box attack} The literature on black-box attacks on graphs remains relatively underexplored. Some existing works focus on poisoning attacks \citep{poisonchang2020restricted}. Other studies, such as \citet{waniek2018hiding} and \citet{xu2019topology}, propose heuristic attacks based on the graph’s topology, but their performance is significantly lower than that of white-box methods like PRBCD. \citet{mu2021hard} approximate gradients by measuring changes with small perturbations, but even under ideal conditions, their method can at best match the performance of PRBCD, which directly utilizes exact gradients. Furthermore, their approach does not scale well to graphs with even a few thousand nodes.

\subsection{Randomized smoothing-based certificates}
\label{app:smoothing_certificates}
A robustness certificate guarantees that the prediction of the classifier remains the same within a specified threat model. For any black-box model, one way to obtain such a guarantee is through randomized smoothing. A smoothing scheme $\xi$ is a random function mapping an input $\vx$ to a nearby point $\vx'$ (e.g. additive isotropic Gaussian noise $\vx’ = \xi(\vx) = \vx + \vepsilon$, where $\vepsilon \sim \mathcal{N}(\boldsymbol{0}, \sigma^2\mI)$ for images). The smooth classifier is defined as the convolution of the smoothing scheme and the black-box classifier $g(\vx) = \Pr[f(\vx + \vepsilon) = y]$ -- majority vote or the probability that the classifier predicts the top class for randomized $\vx' \sim \xi(\vx)$. Regardless of the baseline classifier $f$, smooth classifier $g$ changes slowly around $\vx$ and allows us to bound the worst-case minimum of the smooth prediction probability within $\gB$. For a radius around $\vx$ in which the minimum $g(\tilde\vx)$ remains above $0.5$, we can certify that the smooth model returns the same label (see \autoref{sec:technical} for further details).
In many smoothing schemes, exact computation of the smooth classifier is intractable. The probabilistic computation of it is also expensive as it involves many Monte-Carlo (MC) samples and later accounting for finite sample correction.
\subsection{Conformal prediction}
\label{app: conformal prediction}
Instead of label prediction, conformal prediction (CP) returns prediction sets that are guaranteed to include the true label with adjustable $1 - \alpha$ probability. This post-hoc method treats the model as a black-box and requires only a calibration set of labeled points whose labels were not used during model's training. CP is applicable in both inductive and transductive Graph Neural Networks (GNNs) under the assumption of node-exchangeability  \citep{zargarbashi2024conformal}. 
To compute prediction sets we need to compute a quantile from the set of true calibration conformity scores and compare the scores (e.g. softmaxes) of the test node to the quantile threshold. For i.i.d. data (e.g. images), after computation of the quantile, the task of decreasing the softmax score towards 0 aligns with the goal of decreasing the same value below a conformal threshold (which is by definition above 0). In graphs however this task is more complicated since calibration and test nodes communicate with message passing. \looseness=-1
\section{Issues with Gradient-based Methods}
\label{app: issues_grad}

To motivate the introduction of a search-based method, we first need to understand the shortcomings of using gradients for optimizing the discrete space of the adjacency matrix. We highlight two main problems with gradient-based methods. First, the gradient is a local measure, since it quantifies the behavior of the function under infinitesimal changes. However, we are interested in the behavior of the function when flipping an edge in the discrete space $\{0,1\}$, e.g. from $0$ to $1$. So, flipping an edge could increase the loss even though the gradient suggests that the loss would decrease (and the other way around). This issue was also discussed and illustrated in \citet{zugner2018adversarial} (see their Fig. 4).
Second, even if we assume that the gradient correctly indicates the effect on the loss, it still only reflects the impact of individual changes and ignores the effects of interactions between edges. There are cases where flipping each individual edge would suggest a certain direction of change in the loss (e.g. increase), but flipping both edges together would reverse the direction (e.g. decrease).

\begin{figure}
    \centering
    \includegraphics[width=\linewidth]{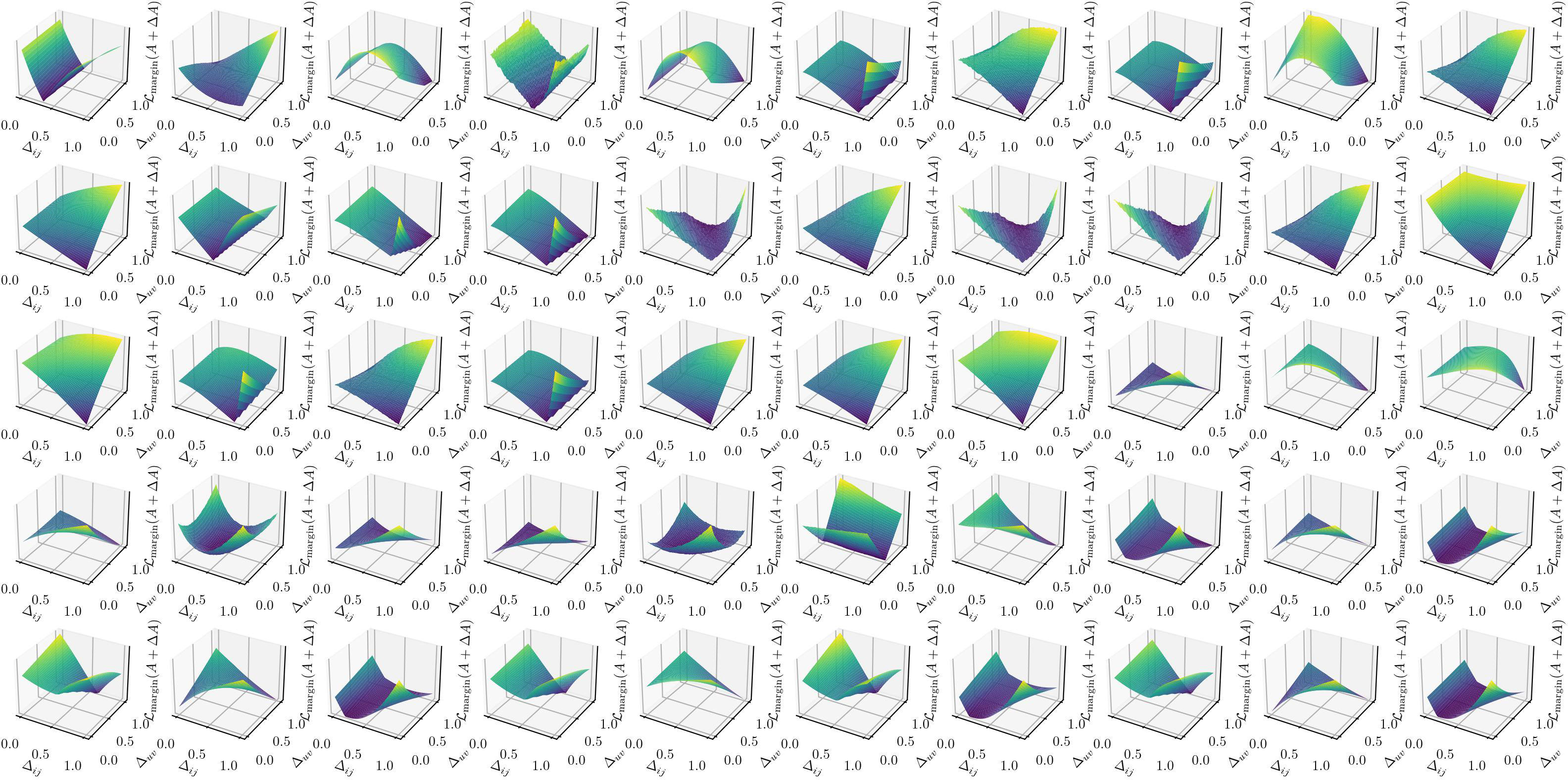}
    \caption{In some cases the gradient fails to measure the effect of flipping an edge on the \texttt{Tanh-Margin} loss. Flipping edges individually vs. jointly has a different effect on loss.}
    \label{fig:grad_problem}
\end{figure}

\begin{figure}[h]
    \centering
    \includegraphics[width=\linewidth]{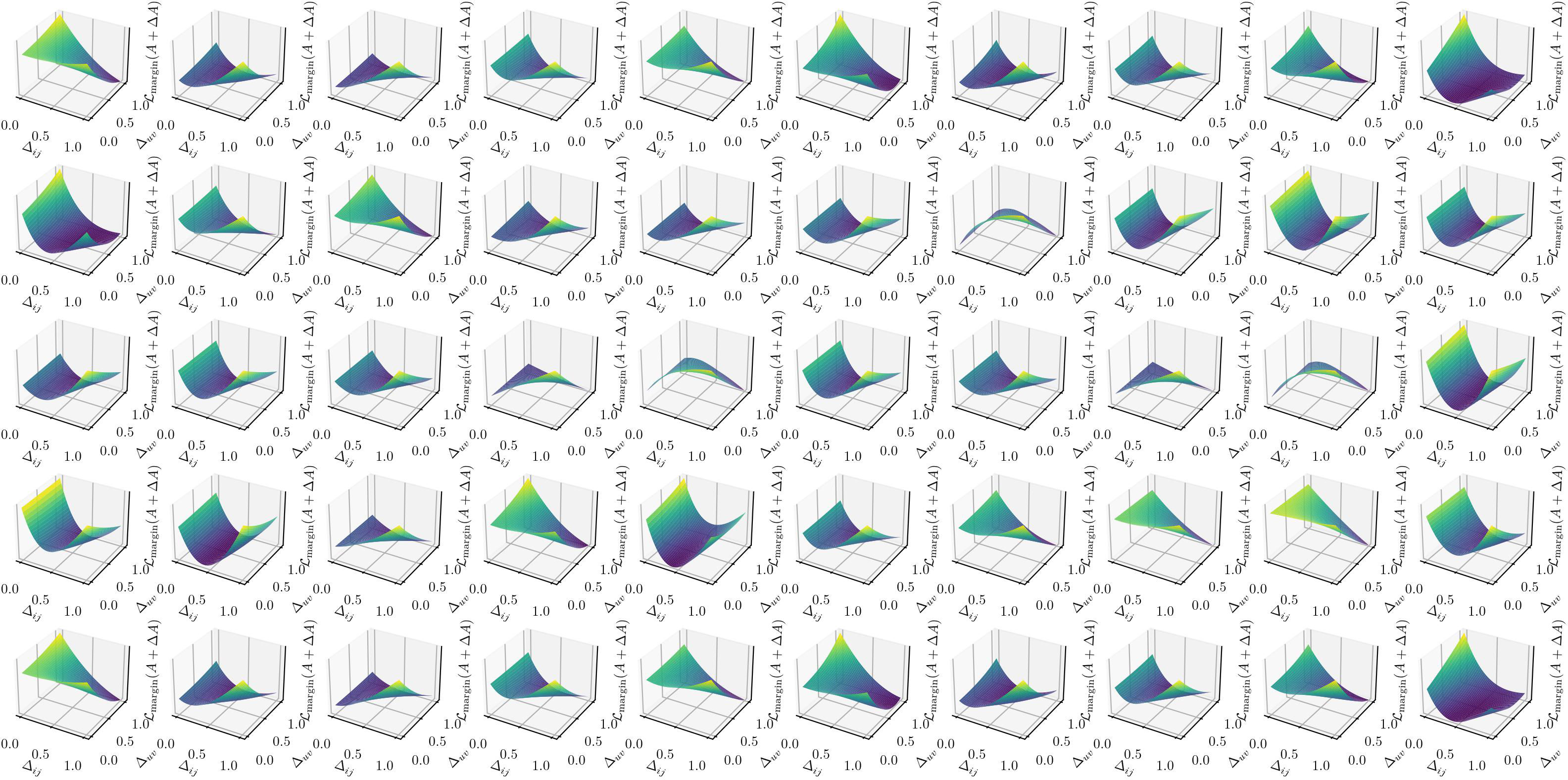}
    \caption{In some cases the gradient fails to measure the effect of flipping and edge on the \texttt{Cross-Entropy} loss. Flipping edges individually vs. jointly has a different effect on loss.}
    \label{fig:graph_problem_ce}
\end{figure}

We designed an experiment to demonstrate that these phenomena are not rare. Since the search space is very large, we start with a specific node and then randomly sample towards the other side of its edges. Specifically, we are looking for the edges $(i,j)$ and $(i, v)$ with $u=i$. We chose this approach because it ensures that the changed edge remains within the first-hop neighborhood of the node. Since our GCN is a two-layer network, the probability that this edge interacts with the two-hop neighborhood of the graph also increases. We found several cases of these two events for each node in the \coraml dataset. \autoref{fig:grad_problem} visualizes a random subset of these cases based on \texttt{Tanh-Margin} loss. The same phenomena also occurs with \texttt{Cross-Entropy} loss. \autoref{fig:graph_problem_ce} visualize a random subset of these cases using  \texttt{Cross-Entropy} loss.

\section{Supplementary Experiments}
\label{sec:experiments-appendix}

\parbold{Transductive Setting}
\label{{app: transductive_setting}}
In \autoref{sec:emp_res}, we argued that transductive setup carreis a false sence of robustness. In this setup, trivial robustness can be gained just by memorization of the clean graph \citep{lrbcd}. For completeness, here we report the results in the transductive setup as well. As in \autoref{tab:coraml_transductive} \method outperforms SOTA consistent with other experiments in the inductive setup.

\input{tables/transductive-table}

\parbold{Stratified sampling}
Although unrealistic, in \autoref{tab:coraml_inductive_stratified}, we compare attacks in case the models are trained train/val/test sampled with the same number of nodes across different classes. Consistent with other results, \method shows to be better here as well. 
\input{tables/stratified_table}

\subsection{Inductive setting, non-stratified sampling}
\label{sec:inductive-expr}
\parbold{Vanilla models}
Here, we present additional results specifically for the inductive setting. With the discussion in \autoref{sec:emp_res} our main experimental setup is for inductive GNNs.  In this section, we detail the effectiveness of our method compared to other approaches. We show the results for \coraml, \pubmed, and \citeseer datasets and vanilla models in tables \ref{tab:coraml-model}, \ref{tab:coraml-attacks}, \ref{tab:pubmed-model}, \ref{tab:pubmed-attack}, \ref{tab:citeseer-model}, and \ref{tab:citeseer-attack}. For each dataset we compare our attack with SOTA on four models: GCN, GAT, APPNP, and GPRGNN. We further compare the local variant of \method with LRBCD under local-degree constraint for GCN, and GPRGNN.

\input{tables/coraml_models}
\input{tables/coraml_attacks}
\input{tables/pubmed_models}
\input{tables/pubmed_attacks}
\input{tables/citeseer_models}
\input{tables/citeseer_attacks}

\parbold{Models with robust training}
\autoref{tab:coraml_inductive} compares \method and SOTA for models training with robust training. During training of these models, we use an adversarial attack at each step to attack $\Gtr$, and then we retrain the model on the adversarially perturbed graph. The robust budget ($\epsilon_\mathrm{robust}$) for adversarial attack during training was $0.2$. This process repeats in each epoch of training until the model converges. \citet{gosch2023adversarial} shows that models with adversarial and self training carry a false sense of robustness in transductive setup, therefore same as other experiments we evaluate in inductive setup. Similar to vanilla models, \method outperforms all previous attacks.

\input{tables/results_table}

\subsection{\arxiv dataset} 

To show the scalability of our attack on larger graphs, we present results on the large \arxiv dataset. We compare \method and PRBCD on the same setup as other experiments. However we propose two other setups where the attack is more realistic: \begin{enumerate*}[label=(\roman*)]
    \item Smaller perturbation budgets: perturbing the \arxiv dataset with the same budget as an smaller graph like \coraml is unrealistic. Therefore we can decrease $\epsilon$, by one order of magnitude and evaluate both methods on  $\epsilon \in \{0.1\%, 0.5\%, 1.0\%\}$. The results for these budgets are summarized in \autoref{table: arxiv_1}.
    \item Simialrly another realistic setup is that on a large graph, the adversary can access a smaller subset of control nodes (e.g. 1000 nodes) with the objective to perturb a set of target nodes. As an example in a social network, an attacker could purchase 1,000 user accounts and use them to influence the performance of other subgroups. Here, we randomly sampled 1,000 nodes as control and 1,500 nodes as target for 5 rounds. We compared \method and PRBCD and reported the average results in \autoref{table: arxiv_2}. Our method outperforms PRBCD in this scenario as well. 
\end{enumerate*}

\begin{minipage}{0.5\textwidth}
\centering
\captionof{table}{Comparison of classification accuracy (\%) on the \texttt{CoraML} dataset under EvA and PRBCD across varying perturbation budgets $\epsilon$.}
\label{tab:performance_comparison}
\vskip 0.5em
\begin{tabular}{lcccc}
\toprule
\textbf{Attack} & \textbf{Clean} & \textbf{0.1\%} & \textbf{0.5\%} & \textbf{1\%} \\ \midrule
PRBCD & 70.53 & 69.83 & 68.64 & 66.27 \\ 
EvA   & 70.53 & 69.21 & 67.59 & 66.86 \\ \bottomrule
\end{tabular}
\label{table: arxiv_1}
\end{minipage}
\begin{minipage}{0.49\textwidth}
\centering
\captionof{table}{Comparison of classification accuracy (\%) on the \texttt{CoraML} dataset under EvA and PRBCD across varying perturbation budgets $\epsilon$ using control nodes.}
\vskip 0.5em
\label{tab:performance_comparison2}
\begin{tabular}{lccc}
\toprule
\textbf{Attack} & \textbf{Clean} & \textbf{1\%} & \textbf{5\%} \\ \midrule
PRBCD &   &     64.89    & 54.7 
\\ 
EvA   &   &  59.3     & 53.92       \\ \bottomrule
\end{tabular}
\label{table: arxiv_2}
\end{minipage}

In addition to both realistic cases, we compared \method (with divide and conqure) to PRBCD in the same setup as we evaluated for other datasets. The summarized result is illusterated in \autoref{fig:all-vanilla-gcn}.

\subsection{Comparison with \citep{rl_attack}}
\label{app: comp: prev_gen_att}

\citep{rl_attack} proposed a practical black-box attack (PBA), dividing it into PBA-C (with access to logits - continuous) and PBA-D (access only to the labels - discrete). As stated in \citep{rl_attack}, a genetic algorithm for global attacks requires PBA-C because it relies on logits, with the fitness function being the negative log-likelihood. We demonstrate that EvA not only eliminates the need for logits but also performs even better by directly optimizing for accuracy rather than using log-likelihood. To compare our method with \citep{rl_attack}, we modified the algorithm's fitness function and mutation mechanism to replicate the results reported in \citep{rl_attack}. This implementation retains scalability benefits, as it is also built upon our sparse encoded representation. Note here we re-implement \citet{rl_attack} in our sparse and parallelized framework. Their \href{https://github.com/Hanjun-Dai/graph_adversarial_attack/tree/master}{original implementation} uses dense adjacency matrices and sequential evaluation and would achieve a significantly worse result within the same memory/run-time constraint. Even with our efficient re-implementation \citet{rl_attack} is significantly worse than ours. \autoref{tab:compare_previous} provides the results for the \coraml dataset using the GCN architecture. \method also significantly outperforms \citep{rl_attack}.
 
Additionally, since our method is independent of gradients, we established the first attack on conformal prediction and certification. For conformal prediction, we attack coverage and set size where the latter criteria are not yet explored (to the best of our knowledge). Attacks tending to decrease certificate effectiveness are also under-explored in GNNs. In this work, we aim to achieve both attack on certified accuracy and certified ratio.

\begin{table}[h!]
\centering
\setlength{\tabcolsep}{3pt} 
\caption{Comparison of classification accuracy (\%) on the \texttt{CoraML} dataset under EvA and \citep{rl_attack} across varying perturbation budgets $\epsilon$.}

\vskip 0.5em

\renewcommand{\arraystretch}{1.4} 
\setlength{\tabcolsep}{2pt} 
\begin{tabular}{lccccccc}
\hline
\textbf{Attack} & \textbf{Clean} & \boldmath$0.01$ & \boldmath$0.02$ & \boldmath$ 0.05$ & \boldmath$0.1$ & \boldmath$ 0.15$ & \boldmath$0.2$ \\ \hline
\citep{rl_attack}& 
$81.07_{\pm{2.07}}$ & 
$78.50_{\pm{1.66}}$ & 
$76.66_{\pm{2.22}}$ & 
$72.53_{\pm{1.91}}$ & 
$68.75_{\pm{1.45}}$ & 
$65.34_{\pm{1.20}}$ & 
$63.27_{\pm{2.47}}$ \\ 
EvA & 
${81.07}_{\pm{2.07}}$ & 
$\textbf{74.80}_{\pm{1.50}}$ & 
$\textbf{68.97}_{\pm{1.58}}$ & 
$\textbf{52.95}_{\pm{1.91}}$ & 
$\textbf{41.99}_{\pm{2.06}}$ & 
$\textbf{37.65}_{\pm{2.74}}$ & 
$\textbf{35.37}_{\pm{2.38}}$ \\ \hline
\end{tabular}
\label{tab:compare_previous}
\end{table}

\section{Technical Details of \method}
\label{sec:technical}
\parbold{Mapping function: enumeration over $\mA$}
For enumerating over $\mA$, instead of using the row and column indices of the node to select, we introduced indexing. For a directed graph, the indexing starts from $0$ to $n^2 - 1$. However, in an undirected graph, we only need the upper triangular part of the matrix $\mA$. To achieve this, we use the following algebraic solution to find the row and column of the perturbation by referencing only the upper triangular indexing.
\begin{equation}
\begin{aligned}
r &= n - 2 - \left\lfloor \frac{\sqrt{-8 l + 4 n(n - 1) - 7}}{2} - 0.5 \right\rfloor \\
c &= 1 + l + r - \frac{n(n - 1)}{2} + \left\lfloor \frac{(n - r)(n - r - 1)}{2} \right\rfloor
\end{aligned}
\end{equation}
The advantage of this solution is that it can also be implemented in a vectorized way, making everything parallelizable.

\parbold{Attacking robustness certificates}
We define a randomized model as a convolution of the original model and a smoothing scheme. Namely the procedure or smooth inference is to add a noise (defined by the smoothing scheme) to the input, and evalute the model on the noisy input. The output of the smooth classifier is the probability of the top class over realizations of the noise (this is the output of the smooth classifier binary certificate; for confidence certificate the output is the expected softmax scores). The smoothing scheme $\xi: \gX \mapsto \gX$ is a randomized function mapping the given input to a random nearby point. For graph structure, we use the sparse smoothing certificate \citep{bojchevski2020efficient}, which certifies whether within $\gB_{r_a, r_d}$ the prediction of the smooth model remains the same. Here $r_a$ is the maximum number of possible additions, and $r_d$ is the maximum number of edge deletions. The smoothing function is defined by two Bernoulli parameters $p_+$, and $p_-$; i.e. for each entity of $\mA$, if it is zero, it will be toggled with $p_+$ probability and otherwise with $p_-$. The same smoothing scheme (and threat model) can be defined for features if the feature space is also binary and sparse. Setting $p_+=p_-$ reduces the certificate to uniform smoothing certificate for $\ell_1$ ball. 

Smoothing certificates require black-box access to the model $f$. As described above the smooth classifier is defined as $\bar{f}_y(\vx) = \E[\sI[f(\xi(\vx)) = y]]$ - each random sample $\vx'$ is one vote for class $f(\vx')$ and $\bar{f}_y$ is the proportion of votes for class $y$. Regardless of the model $f$, the smooth model $\bar{f}$ changes slowly around the input.
Let $p = \bar{f}_y(\vx)$; for the smooth classifier we can find a lower bound probability $\underline{p} \ge \min_{\tilde\vx \in\gB(\vx)} \bar{f}_y(\tilde\vx)$ and define the certificate as a decision function $\sI[\underline{y} \ge 0.5]$. This decision function ensures that the predicted class still remains the top-class for any point within the threat model. For details including the optimization function and how to compute certified lower bound see \citep{bojchevski2020efficient}.  

Whether a node (an input in general) is certified reduces to whether the smooth prediction probability for the input is above a threshold $\bar{p}$. This is due to the non-decreasing property of the certified ratio with respect to $\bar{p}$. Additionally since the certificate is only a function of the probability and not the input, we can find this value easily via binary search. Therefore our objective is to decrease the probability of the smooth classifier below $\bar{p}$ for as many node as possible.

\parbold{Adaptive sampling for certificate attack} Statistical rigor is not a necessity while attacking the certificate. Therefore, we can reduce the sampling rate to a low number while finding the perturbation. Later to ensure that our attack has reduces the certified ratio we again follow the proper certificate configuration. During the attack, we can reduce the cost of resampling by only resampling the subset of the graph that was perturbed. In other words, we initialize the search by computing and storing samples $\mA_1, \dots, \mA_m$ from the clean graph, and for each perturbation $\tilde\mA$ we only need to resample the edges in $\mA \triangle \tilde\mA$. Specifically for any edge removed from the graph during perturbation we update original sample $\mA_1, \dots, \mA_m$ with $p_+$ Bernoulli samples in the same index of the added edge.  Similar process is done with $p_-$ random edge removals for edges added in the perturbation. We substitute those samples in the same entry of $\mA_1, \dots, \mA_m$, and by running this process $|\delta|$ times, we assume that $\tilde\mA_1, \cdots \tilde\mA_m$ are representative as a new set of $m$ samples for $\tilde\mA$. This adaptive sampling reduces the number of random computations from $m\cdot n^2$ to $m\cdot|\delta|$, which is significantly lower. Surely, to evaluate the final perturbation (the reported effectiveness), we don't use this approach, as it is statistically flawed and only applicable to reduce the computation during the attack.

\subsection{Label diversity}  
\label{app:labeldiversity}
\begin{figure*}[b!]
    \centering
    \input{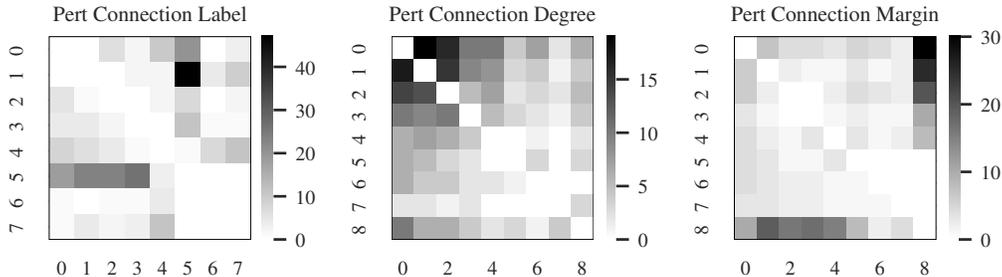}
    \caption{The upper triangle of each heatmap represents the perturbation connections for PRBCD, the lower triangle corresponds to the same for \method, and the diagonal is set to zero.}
    \label{fig: abl_pert_connection}
\end{figure*}

We further conduct an ablation study on the solutions found by \method and PRBCD under a specific budget of 10\%. In this experiment, we keep all hyperparameters of \method and PRBCD fixed and run them across 10 different seeds. We then compare the average solutions generated by each adversary.
The left figure in \autoref{fig: abl_pert_connection} shows the number of connections across different labels. In both cases, the methods focus more on label 5 than on the others, but \method distributes the connections more uniformly compared to PRBCD. The middle figure illustrates the nodes with original degrees ranging from 1 to greater than 8. The results indicate that, in both attacks, most of the budget is spent connecting to low-degree nodes. However, compared to PRBCD, \method allocates more of the budget to higher-degree nodes.
Additionally, we calculate the margin loss for each node in the original graph and discretize them into eight levels. As shown in the right figure of \autoref{fig: abl_pert_connection}, \method allocates more of the budget to higher-margin nodes, resulting in a non-trivial solution that achieves a better optimum. Finally, it seems that \method identifies solutions that differ from greedy-based heuristic, which usually only targets low-degree or low-margin nodes.

\subsection{Time Analysis}
We run an ablation study comparing PRBCD, and \method for wall clock time and memory. In \method, the number of steps controls the time and the size of the population (assuming all population is evaluated at once using stacked inference) controls the memory. Similarly for PRBCD, time is controlled by the number of epochs controls the time and memory is a function of block size. We evaluate \method with different numbers of steps, population sizes, and parallel evaluations, and PRBCD with varying numbers of epochs and block sizes on the \pubmed dataset. \autoref{fig: scaling-time} (left) shows the results for \method and PRBCD in terms of memory usage, wall clock time and method performance. Our method demonstrates comparable performance within the same level of wall clock time (less than a minute). Moreover, by increasing the wall clock time—through and memory either by a larger population size or more steps— \method enhances in performance. This is while PRBCD an almost constant trend given more time or memory. 

\begin{figure}
    \centering
    \input{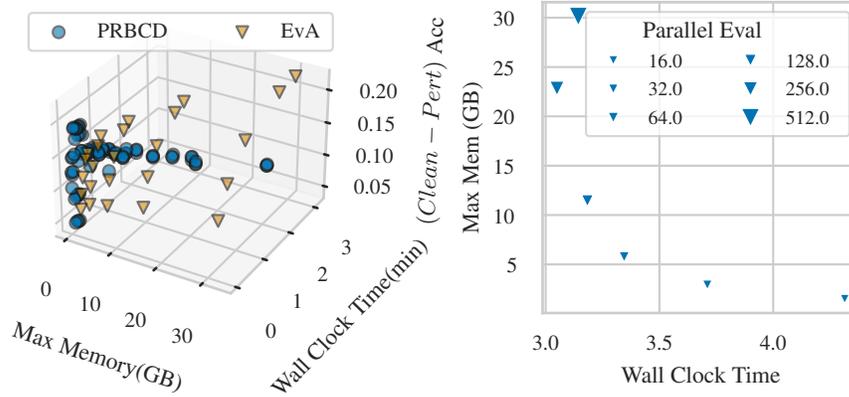}
    \caption{Comparing the memory usage between \method and PRBCD.}
    \label{fig: time_analysis}
\end{figure}

Additionally, in \autoref{fig: time_analysis}, we highlight how our framework provides a trade-off between time and memory for achieving the same level of accuracy by varying the number of parallel evaluations. For each point in the figure, we observe roughly the same performance; 
however, the methods differ in memory usage due to different number of parallel evaluation, leading to variations in wall clock time.

\subsection{Divide and Conquer}
\label{sec:d-and-c}

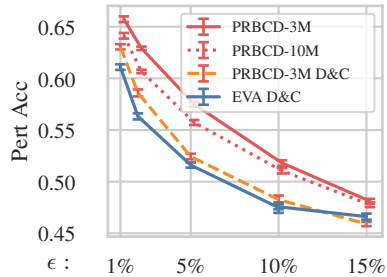
\begin{figure}[h]
    \centering
    \input{./figures/dc_scaling.pgf}
    \caption{Effect of Divide on Conquer on \method and PRBCD.}
    \label{fig:divide-conqure}
\end{figure}

Computing gradients w.r.t. all elements in matrix $\mA$ is computationally intensive. And as the graph grows the elements for which we should compute and store gradient increases quadratically. As a remedy PRBCD applies a block-coordinate gradient descent where in each iteration the gradients are computed over a subset of indices in $\mA$. Intuitively PRBCD works under this assumption that a relaxation from $\mA$ to a (random) subset of adjacency matrix does change the optimal solution by a high margin. In \autoref{fig:divide-conqure} for \arxiv dataset, we increased the block size of PRBCD, from the suggested 3M to 10M, and still the result is far from \method. Beyond that block size could not fit into the memory. 

While \method works with sparse representation of $\mA$ still the search space ($2^{|\mA|}$) grows exponentially with the number of nodes. Via divide and conquer (D\&C) we apply relaxation where we assume a sequential search for optimal attack targeting disjoint subsets of $\Vattack$ does not result far from the optimal solution for the entire $\Vattack$. 
Therefore we divide this set into $k$ disjoint subsets and run the attack over each subset separately. Our attack applies on subsets in a sequential order meaning that after attacking one subset, the attacked graph is assumed to be the clean baseline to attack the other set. At the final round, the attack carries all the perturbations applied on $\Vattack$. To ensure the validity of the result, we divide the budget according to the edges connected to each subset. 

As our D\&C approach is applicable regardless of the attack algorithm we can similarly apply it to PRBCD. Although it does not outperform \method, we show that adding D\&C to PRBCD increases the performance by a significant margin.

\section{Datasets and Models, and Hyperparameters}
\label{sec:dataset-models-hyperparameters}

\subsection{Statistics of datasets}
In our experiments, we mainly conduct experiments on the commonly used graph datasets: \coraml, \citeseer, and \pubmed, which are all representative academic citation networks. Their specific characteristics are as follows:

\parbold{\coraml} The \coraml dataset contains 2,810 papers as nodes, with citation relationships between them as edges, resulting in 7,981 edges. Each paper is categorized into one of 7 classes corresponding to different subfields of machine learning. Each node is represented by a 1,433-dimensional bag-of-words (BoW) feature vector derived from the words in the titles and abstracts of the papers.

\parbold{\citeseer} The \citeseer dataset is also an academic citation network dataset consisting of 3,312 papers from 6 subfields of computer science and a total of 4,732 citation edges. Similar to \coraml, each paper as a node is represented by a BoW feature vector with a dimensionality of 3,703.

\parbold{\pubmed} The \pubmed dataset is derived from a citation network of biomedical literature that contains 19,717 papers as nodes and 44,338 citation edges. Each paper is categorized into one of 3 classes based on its topic. The node features in \pubmed are 500-dimensional vectors.

\parbold{\computers\  and \photo} The \computers\  and \photo\  datasets consists of two networks of Amazon Computers and Amazon Photo. In these networks, nodes represent individual goods sold on Amazon, and edges indicate that two products are frequently purchased together. Each node is accompanied by bag-of-words features derived from product reviews, providing a textual representation of the item's description and customer feedback. The task is predicting the product category.

\begin{table}[htbp]
\centering
\caption{Dataset statistics.}
\vskip 0.5em
\label{tab:dataset_stats}
\begin{tabular}{lcccc}
\toprule
\textbf{Dataset}   & \textbf{Nodes} & \textbf{Edges} & \textbf{Features} & \textbf{Classes} \\ \midrule
\textbf{\coraml}   & 2,810          & 7,981          & 1,433             & 7                \\
\textbf{\citeseer}  & 3,312          & 4,732          & 3,703             & 6                \\
\textbf{\texttt{\pubmed}}    & 19,717         & 44,338         & 500               & 3                \\ 
\textbf{\computers}    & 13,752         & 491,722         & 767               & 10                \\ \textbf{\photo}    & 7,650         & 238,162
         & 745              & 8             
\\
\bottomrule
\end{tabular}
\end{table}

\subsection{Details of models}
In the following sections, we detail the hyperparameters and architectural details for the models performed in this paper. 

\parbold{GCN} We utilize a two-layer GCN with 64 hidden units. A dropout rate of 0.5 is applied during training.

\parbold{GAT} Our GAT model consists of two layers with 64 hidden units and a single attention head. During training, we apply a dropout rate of 0.5 to the hidden units, but no dropout is applied to the neighborhood.

\parbold{APPNP}  We use a two-layer MLP with 64 hidden units to encode the node attributes. We then apply generalized graph diffusion, using a transition matrix and coefficients $\gamma_{K}=(1-\alpha)K$ and $\gamma_{l}=\alpha(1-\alpha)l$ for $l<K$.

\parbold{GPRGNN} Similar to APPNP, we employ a two-layer MLP with 64 hidden units for the predictive part. We use the symmetric normalized adjacency matrix with self-loops as the transition matrix and randomly initialize the diffusion coefficients. We consider a total of $K = 10$ diffusion steps, with $\alpha$ set to 0.1. During training, we apply a dropout rate of 0.2 to the MLP, while no dropout is applied to the adjacency matrix. Unlike the method in \cite{chien2021adaptive}, we always learn the diffusion coefficients with weight decay, which acts as a regularization mechanism to prevent the coefficients from growing indefinitely.

\begin{table}[t]
    \centering
    \caption{Hyper-parameters for PRBCD, LRBCD, and \method.}
    \vskip 0.5em
    \label{tab:combined-hyperparams}
    \begin{tabular}{lcc|lc}
        \toprule
        \textbf{Hyper-parameter} & \textbf{PRBCD} & \textbf{LRBCD} & \textbf{Hyper-parameter} & \textbf{\method} \\
        \midrule
        Epochs & 500 & 500 & No.\ Steps & 500 \\
        Fine-tune Epochs & 100 & 0 & Mutation Rate & 0.01 \\
        Keep Heuristic & WeightOnly & WeightOnly & Tournament Size & 2 \\
        Search Space Size & 500,000 & 500,000 & Population Size & 1,024 \\
        Loss Type & tanhMargin & $\tanh$-Margin & No.\ Crossovers & 30 \\
        Early Stopping & N/A & False & Mutation Method & Adaptive \\
        \bottomrule
    \end{tabular}
\end{table}

\parbold{SoftMedian GDC} We follow the default configuration from \cite{geisler2023robustnessgraphneuralnetworks}, using a temperature of $T = 0.2$ or the SoftMedian aggregation, with 64 hidden dimensions and a dropout rate of 0.5. We fix the Personalized PageRank diffusion coefficient to $\alpha = 0.15$ and apply a top $k = 64$ sparsification. During the attacks, the model remains fully differentiable, except for the sparsification of the propagation matrix.

\parbold{MLP} We design the MLP following the prediction module of GPRGNN and APPNP, incorporating two layers with 64 hidden units. During training, we apply a dropout rate of 0.2 to the hidden layer.

\subsection{Hyperparameter setup}
In \method we set the capacity of the computation to the same as the population, this means that all perturbations within a population are in one combined inference. However, in some cases where the graph is large (e.g. \pubmed), we reduce this number.

\autoref{tab:combined-hyperparams} shows the hyper-parameter selection in almost all experiments. We only change the population number in some experiments, like certificate attacks, to reduce the computation. E.g., in the certificate attack, the population is reduced by a factor of 10. Finally, all of the experiments has been run in one Nvidia H100 gpu.

\subsection{Attack hyperparameters}
\label{attack-hyperparameters}
To assess the robustness of GNNs, we utilize the following attacks and hyperparameters. Based on \cite{geisler2023robustnessgraphneuralnetworks}, we also select the tanh-margin loss as the attack objective.

\parbold{PRBCD} We closely adhere to the setup outlined by \cite{geisler2023robustnessgraphneuralnetworks}. A block size of 500,000 is used with 500 training epochs. Afterward, the model state from the best epoch is restored, followed by 100 additional epochs with a decaying learning rate and no block resampling. Additionally, the learning rate is scaled according to $\delta$ and the block size, as recommended by \cite{geisler2023robustnessgraphneuralnetworks}.

\parbold{LRBCD} The same block size of 500,000 is used with 500 training epochs. The learning rate is scaled based on $\delta$ and the block size, following the same approach as PRBCD. The local budget is consistently set as 0.5.

\parbold{EvA} We set the population size to 1024 in most cases. Our mutation rate is 0.01, and increasing this number breaks the balance between exploration and exploitation, leading to less effective attacks. We run each attack for 500 iterations in most cases. In cases like certificate attacks, which are time-consuming, we reduce this number to 100. The details are summarized in \autoref{tab:combined-hyperparams}.

For \arxiv dataset we divide the $\Vattack$ to $k_\mathrm{dc} = 98$ subsets where each division includes 500 vertices. There we set the population size to $45$ candidates. The $\delta_i$ for subset $i$ is set to $\epsilon_i = \epsilon \cdot |\gE[\gV_i: \gV|$. We also reduce the number of iterations for \method to 300 for each subset while for PRBCD this number remains 500.

\parbold{EvA-Local} All hyper-parameters are the same as \method. An additional hyper parameter is $t_\mathrm{warm}$ which is the number of initial steps where the random local projection is applied instead of the frequency score-based local projection. This number is set to 50 for \pubmed dataset. Interestingly even without this projection, EvALocal outperforms LRBCD for \citeseer and \coraml datasets. There we only remove random matchings from the nodes with degree violation until the total violation reaches 0. This approach does not work on the \pubmed dataset. The intuition is that since \pubmed larger and more dense compared to other datasets, for each candidate there are a lot of edges that has at least one endpoint violating the local constraint. Therefore removing many edges from a candidate, and replacing them by random edge adds a large random noise to each condidate at each iteration. Therefore the search is done over a very noisy setup.

\parbold{PGA} For the PGA, we adopt the same setting as in \cite{zhu2023simpleefficientpartialgraph}. We use GCN as the surrogate model and tanhMarginMCE-0.5 as the loss type. The attack is configured with 1 greedy step, a pre-selection ratio of 0.1, and a selection ratio of 0.6. Additionally, the influence ratio is set to 0.8, with the selection policy based on node degree and margin.

\end{document}

%% file: math_commands.tex
\usepackage{amsmath,amsfonts,bm}

\def\eqref#1{equation~\ref{#1}}

\def\1{\bm{1}}

\def\ve{{\bm{e}}}

\def\vs{{\bm{s}}}

\def\vx{{\bm{x}}}
\def\vy{{\bm{y}}}
\def\vz{{\bm{z}}}

\def\mA{{\bm{A}}}

\def\mI{{\bm{I}}}

\def\mP{{\bm{P}}}

\def\mX{{\bm{X}}}

\DeclareMathAlphabet{\mathsfit}{\encodingdefault}{\sfdefault}{m}{sl}
\SetMathAlphabet{\mathsfit}{bold}{\encodingdefault}{\sfdefault}{bx}{n}

\def\gA{{\mathcal{A}}}
\def\gB{{\mathcal{B}}}

\def\gE{{\mathcal{E}}}

\def\gG{{\mathcal{G}}}

\def\gL{{\mathcal{L}}}

\def\gO{{\mathcal{O}}}

\def\gS{{\mathcal{S}}}

\def\gV{{\mathcal{V}}}

\def\gX{{\mathcal{X}}}

\def\sI{{\mathbb{I}}}

\def\sR{{\mathbb{R}}}

\newcommand{\E}{\mathbb{E}}

\DeclareMathOperator*{\argmax}{arg\,max}

%% file: tables/transductive-table.tex

\begin{table}[ht]
    \centering
    \setlength{\tabcolsep}{3pt} 
    \caption{Classification accuracy (\%) on the \texttt{CoraML} dataset in the transductive setting under adversarial attacks. 
    Results are reported for two GNN models subjected to three different attack methods across varying perturbation budgets $\epsilon$. 
    Training, validation, and test sets are non-stratified.}
    \vskip 0.5em
    \label{tab:coraml_transductive}
    \begin{tabular}{llcccccc}
    \toprule
    \multirow{2}{*}{Model} & \multirow{2}{*}{Attack} & \multicolumn{6}{c}{$\epsilon$} \\
    \cmidrule(lr){3-8}
     & & 0.01 & 0.02 & 0.05 & 0.10 & 0.15 & 0.20 \\
    \midrule
    \multirow{3}{*}{GCN} 
       & LRBCD & $79.88_{\pm 1.52}$ & $77.66_{\pm 1.84}$ & $72.61_{\pm 1.38}$ & $65.84_{\pm 1.44}$ & $60.80_{\pm 1.82}$ & $56.86_{\pm 1.92}$ \\
       & PRBCD & $79.48_{\pm 1.70}$ & $77.26_{\pm 1.59}$ & $71.85_{\pm 1.80}$ & $65.49_{\pm 1.73}$ & $60.49_{\pm 2.29}$ & $56.07_{\pm 2.36}$ \\
       & EvA   & $77.38_{\pm 1.87}$ & $74.12_{\pm 1.99}$ & $65.68_{\pm 1.97}$ & $60.49_{\pm 2.31}$ & $58.72_{\pm 2.57}$ & $57.41_{\pm 2.51}$ \\
    \midrule
    \multirow{3}{*}{GPRGNN} 
       & LRBCD & $79.24_{\pm 3.04}$ & $77.09_{\pm 3.25}$ & $71.66_{\pm 4.43}$ & $62.74_{\pm 7.28}$ & $55.01_{\pm 11.96}$ & $49.04_{\pm 15.66}$ \\
       & PRBCD & $78.67_{\pm 3.20}$ & $75.86_{\pm 3.69}$ & $69.83_{\pm 5.09}$ & $62.15_{\pm 7.47}$ & $55.15_{\pm 10.17}$ & $50.13_{\pm 12.63}$ \\
       & EvA   & $76.53_{\pm 3.46}$ & $72.74_{\pm 4.53}$ & $63.78_{\pm 6.19}$ & $56.83_{\pm 9.98}$ & $53.00_{\pm 13.01}$ & $50.04_{\pm 15.53}$ \\
    \bottomrule
    \end{tabular}
\end{table}

%% file: tables/stratified_table.tex
\begin{table}[ht]
    \centering
    \setlength{\tabcolsep}{2.5pt} 
    \caption{Classification accuracy (\%) on the \texttt{CoraML} dataset in the inductive setting under adversarial attacks. 
    Results are reported for two GNN models subjected to three different attack methods across varying perturbation budgets $\epsilon$.
    Training, validation, and test sets are stratified.}
    \vskip 0.5em
    \label{tab:coraml_inductive_stratified}
    \begin{tabular}{llcccccc}
    \toprule
    \multirow{2}{*}{Model} & \multirow{2}{*}{Attack} & \multicolumn{6}{c}{$\epsilon$} \\
    \cmidrule(lr){3-8}
     & & 0.01 & 0.02 & 0.05 & 0.10 & 0.15 & 0.20 \\
    \midrule
    \multirow{3}{*}{GCN} 
       & LRBCD & $80.00_{\pm 2.70}$ & $77.43_{\pm 2.58}$ & $71.43_{\pm 2.72}$ & $65.50_{\pm 3.91}$ & $61.21_{\pm 4.25}$ & $57.64_{\pm 4.54}$ \\
       & PRBCD & $78.71_{\pm 2.80}$ & $75.29_{\pm 3.42}$ & $67.86_{\pm 3.43}$ & $59.50_{\pm 3.50}$ & $53.00_{\pm 4.14}$ & $48.43_{\pm 3.73}$ \\
       & EvA   & $77.00_{\pm 2.86}$ & $71.36_{\pm 3.29}$ & $54.36_{\pm 4.73}$ & $44.29_{\pm 3.28}$ & $40.86_{\pm 3.73}$ & $37.29_{\pm 3.73}$ \\
    \midrule
    \multirow{3}{*}{GPRGNN} 
       & LRBCD & $76.21_{\pm 7.94}$ & $73.14_{\pm 7.88}$ & $66.07_{\pm 11.50}$ & $60.79_{\pm 11.74}$ & $56.43_{\pm 12.43}$ & $53.36_{\pm 12.90}$ \\
       & PRBCD & $74.36_{\pm 9.60}$ & $70.71_{\pm 9.96}$ & $63.79_{\pm 10.44}$ & $56.14_{\pm 11.14}$ & $49.29_{\pm 11.44}$ & $45.07_{\pm 11.04}$ \\
       & EvA   & $71.86_{\pm 10.37}$ & $65.00_{\pm 11.44}$ & $50.14_{\pm 11.44}$ & $41.79_{\pm 14.02}$ & $37.21_{\pm 14.58}$ & $35.21_{\pm 15.67}$ \\
    \bottomrule
    \end{tabular}
\end{table}

%% file: tables/coraml_models.tex
\begin{table}[ht]
    \centering
    \setlength{\tabcolsep}{4pt} 
    \caption{Classification accuracy (\%) on the \texttt{CoraML} dataset in the inductive setting under adversarial attacks. 
    Results are reported for four GNN models subjected to two different attack methods across varying perturbation budgets $\epsilon$.
    Training, validation, and test sets are non-stratified.}
    \label{tab:coraml-model}
    \vskip 0.5em
    \begin{tabular}{llcccccc}
    \toprule
    \multirow{2}{*}{Model} & \multirow{2}{*}{Attack} & \multicolumn{6}{c}{$\epsilon$} \\
    \cmidrule(lr){3-8}
    & & 0.01 & 0.02 & 0.05 & 0.10 & 0.15 & 0.20 \\
    \midrule
    \multirow{2}{*}{APPNP} 
        & EvA & $76.65_{\pm 1.32}$ & $71.03_{\pm 1.44}$ & $56.51_{\pm 1.60}$ & $49.32_{\pm 1.84}$ & $44.77_{\pm 2.04}$ & $41.42_{\pm 1.41}$ \\
    & PRBCD & $78.65_{\pm 0.99}$ & $75.30_{\pm 1.27}$ & $68.75_{\pm 1.22}$ & $61.57_{\pm 1.65}$ & $55.44_{\pm 1.58}$ & $49.96_{\pm 2.42}$ \\
    \midrule
    \multirow{2}{*}{GAT} 
        & EvA & $64.20_{\pm 1.89}$ & $58.51_{\pm 2.45}$ & $40.99_{\pm 1.60}$ & $15.30_{\pm 4.47}$ & $9.40_{\pm 6.83}$ & $8.11_{\pm 6.65}$ \\
    & PRBCD & $70.07_{\pm 2.82}$ & $66.55_{\pm 2.21}$ & $58.58_{\pm 3.33}$ & $49.61_{\pm 6.55}$ & $39.86_{\pm 6.78}$ & $36.94_{\pm 7.09}$ \\
    \midrule
    \multirow{2}{*}{GCN} 
        & EvA & $74.80_{\pm 1.50}$ & $68.97_{\pm 1.58}$ & $52.95_{\pm 1.91}$ & $41.99_{\pm 2.06}$ & $37.65_{\pm 2.74}$ & $35.37_{\pm 2.38}$ \\
    & PRBCD & $76.44_{\pm 1.64}$ & $73.17_{\pm 1.39}$ & $66.48_{\pm 2.13}$ & $58.51_{\pm 1.77}$ & $52.67_{\pm 2.09}$ & $47.19_{\pm 2.02}$ \\
    \midrule
    \multirow{2}{*}{GPRGNN} 
        & EvA & $72.53_{\pm 4.11}$ & $66.83_{\pm 4.54}$ & $51.53_{\pm 5.57}$ & $42.21_{\pm 8.52}$ & $37.01_{\pm 9.83}$ & $34.52_{\pm 9.83}$ \\
    & PRBCD & $74.95_{\pm 3.08}$ & $71.67_{\pm 2.76}$ & $64.84_{\pm 4.18}$ & $57.94_{\pm 4.55}$ & $53.24_{\pm 5.20}$ & $48.68_{\pm 6.52}$ \\
    \bottomrule
    \end{tabular}
\end{table}

%% file: tables/coraml_attacks.tex
\begin{table}
    \centering
    \setlength{\tabcolsep}{3pt} 
    \caption{Classification accuracy (\%) on the \texttt{CoraML} dataset in the inductive setting under adversarial attacks. 
    Results are reported for two GNN models subjected to five different attack methods across varying perturbation budgets $\epsilon$.
    Training, validation, and test sets are non-stratified.}
    \label{tab:coraml-attacks}
    \vskip 0.5em
    \begin{tabular}{llcccccc}
    \toprule
    \multirow{2}{*}{Model} & \multirow{2}{*}{Attack} & \multicolumn{6}{c}{$\epsilon$} \\
    \cmidrule(lr){3-8}
    & & 0.01 & 0.02 & 0.05 & 0.10 & 0.15 & 0.20 \\
    \midrule
    \multirow{4}{*}{GCN} 
        & EvA & $74.80_{\pm 1.50}$ & $68.97_{\pm 1.58}$ & $52.95_{\pm 1.91}$ & $41.99_{\pm 2.06}$ & $37.65_{\pm 2.74}$ & $35.37_{\pm 2.38}$ \\
    & EvaLocal & $75.09_{\pm 1.73}$ & $69.82_{\pm 1.96}$ & $60.21_{\pm 2.04}$ & $56.09_{\pm 1.93}$ & $54.16_{\pm 2.48}$ & $52.88_{\pm 1.79}$ \\
    & LRBCD & $78.51_{\pm 1.56}$ & $75.94_{\pm 1.54}$ & $71.10_{\pm 1.16}$ & $64.41_{\pm 1.65}$ & $60.14_{\pm 1.73}$ & $57.37_{\pm 1.45}$ \\
    & PRBCD & $76.44_{\pm 1.64}$ & $73.17_{\pm 1.39}$ & $66.48_{\pm 2.13}$ & $58.51_{\pm 1.77}$ & $52.67_{\pm 2.09}$ & $47.19_{\pm 2.02}$ \\
    & PGA & $79.58_{\pm 1.61}$ & $76.92_{\pm 1.73}$ & $70.94_{\pm 1.89}$ & $64.62_{\pm 1.92}$ & $60.46_{\pm 2.25}$ & $57.54_{\pm 2.46}$ \\
    
    \midrule
    \multirow{4}{*}{GPRGNN} 
        & EvA & $72.53_{\pm 4.11}$ & $66.83_{\pm 4.54}$ & $51.53_{\pm 5.57}$ & $42.21_{\pm 8.52}$ & $37.01_{\pm 9.83}$ & $34.52_{\pm 9.83}$ \\
    & EvaLocal & $73.31_{\pm 3.30}$ & $67.26_{\pm 4.17}$ & $58.29_{\pm 7.96}$ & $53.38_{\pm 11.42}$ & $51.10_{\pm 12.66}$ & $49.96_{\pm 13.63}$ \\
    & LRBCD & $77.51_{\pm 1.81}$ & $74.80_{\pm 1.41}$ & $68.83_{\pm 1.90}$ & $62.56_{\pm 1.71}$ & $59.07_{\pm 1.53}$ & $55.66_{\pm 1.71}$ \\
    & PRBCD & $74.95_{\pm 3.08}$ & $71.67_{\pm 2.76}$ & $64.84_{\pm 4.18}$ & $57.94_{\pm 4.55}$ & $53.24_{\pm 5.20}$ & $48.68_{\pm 6.52}$ \\
    & PGA & $78.55_{\pm 3.03}$ & $75.33_{\pm 3.69}$ & $68.63_{\pm 5.11}$ & $61.55_{\pm 6.97}$ & $56.60_{\pm 8.52}$ & $54.91_{\pm 7.46}$ \\
    
    \bottomrule
    \end{tabular}
\end{table}

%% file: tables/pubmed_models.tex
\begin{table}[ht]
    \centering
    \setlength{\tabcolsep}{4pt} 
    \caption{Classification accuracy (\%) on the \texttt{PubMed} dataset in the inductive setting under adversarial attacks. 
    Results are reported for four GNN models subjected to two different attack methods across varying perturbation budgets $\epsilon$.
    Training, validation, and test sets are non-stratified.}
    \label{tab:pubmed-model}
    \vskip 0.5em
    \begin{tabular}{llcccccc}
    \toprule
    \multirow{2}{*}{Model} & \multirow{2}{*}{Attack} & \multicolumn{6}{c}{$\epsilon$} \\
    \cmidrule(lr){3-8}
    & & 0.01 & 0.02 & 0.05 & 0.10 & 0.15 & 0.20 \\
    \midrule
    \multirow{2}{*}{APPNP} 
    & EvA & $73.85_{\pm 2.35}$ & $69.64_{\pm 2.16}$ & $57.07_{\pm 2.32}$ & $47.03_{\pm 2.18}$ & $43.94_{\pm 1.83}$ & $41.93_{\pm 2.18}$ \\
    & PRBCD & $75.54_{\pm 2.34}$ & $72.44_{\pm 2.28}$ & $65.14_{\pm 2.26}$ & $57.15_{\pm 2.59}$ & $51.04_{\pm 2.79}$ & $45.75_{\pm 2.60}$ \\
    \midrule
    \multirow{2}{*}{GAT}
    & EvA & $69.15_{\pm 1.83}$ & $64.62_{\pm 1.81}$ & $52.17_{\pm 1.71}$ & $33.62_{\pm 2.05}$ & $26.62_{\pm 3.74}$ & $24.21_{\pm 4.27}$ \\
    & PRBCD & $71.33_{\pm 1.53}$ & $67.78_{\pm 1.79}$ & $59.73_{\pm 2.10}$ & $49.87_{\pm 1.36}$ & $42.04_{\pm 1.57}$ & $35.94_{\pm 1.66}$ \\
    \midrule
    \multirow{2}{*}{GCN} 
    & EvA & $72.60_{\pm 2.19}$ & $68.35_{\pm 2.41}$ & $56.15_{\pm 1.92}$ & $42.93_{\pm 2.64}$ & $40.46_{\pm 2.76}$ & $39.11_{\pm 2.98}$ \\
    & PRBCD & $74.99_{\pm 1.99}$ & $71.90_{\pm 2.03}$ & $64.16_{\pm 2.32}$ & $55.54_{\pm 2.79}$ & $49.32_{\pm 2.66}$ & $43.90_{\pm 3.09}$ \\
    \midrule
    \multirow{2}{*}{GPRGNN} 
    & EvA & $72.01_{\pm 4.18}$ & $67.61_{\pm 4.28}$ & $55.95_{\pm 4.32}$ & $51.49_{\pm 7.51}$ & $49.18_{\pm 7.83}$ & $42.39_{\pm 9.63}$ \\
    & PRBCD & $74.37_{\pm 3.40}$ & $71.66_{\pm 3.55}$ & $64.51_{\pm 4.94}$ & $56.21_{\pm 6.46}$ & $50.26_{\pm 7.41}$ & $45.81_{\pm 8.47}$ \\
    \bottomrule
    \end{tabular}
\end{table}

%% file: tables/pubmed_attacks.tex
\begin{table}
    \centering
    \setlength{\tabcolsep}{4pt} 
    \caption{Classification accuracy (\%) on the \texttt{PubMed} dataset in the inductive setting under adversarial attacks. 
    Results are reported for two GNN models subjected to four different attack methods across varying perturbation budgets $\epsilon$.
    Training, validation, and test sets are non-stratified.}
    \vskip 0.5em
    \begin{tabular}{llcccccc}
    \toprule
    \multirow{2}{*}{Model} & \multirow{2}{*}{Attack} & \multicolumn{6}{c}{$\epsilon$} \\
    \cmidrule(lr){3-8}
    & & 0.01 & 0.02 & 0.05 & 0.10 & 0.15 & 0.20 \\
    \midrule
    \multirow{4}{*}{GCN}
    & EvA & $72.60_{\pm 2.18}$ & $68.35_{\pm 2.41}$ & $56.15_{\pm 1.92}$ & $42.93_{\pm 2.64}$ & $40.46_{\pm 2.76}$ & $39.11_{\pm 2.98}$ \\
    & EvaLocal & $74.12_{\pm 2.19}$ & $69.99_{\pm 2.04}$ & $63.43_{\pm 2.76}$ & $61.51_{\pm 2.64}$ & $61.01_{\pm 2.79}$ & $60.54_{\pm 2.64}$ \\
    & LRBCD & $74.89_{\pm 2.04}$ & $71.48_{\pm 2.49}$ & $65.68_{\pm 2.90}$ & $60.24_{\pm 3.15}$ & $56.81_{\pm 3.02}$ & $54.07_{\pm 2.99}$ \\
    & PRBCD & $74.99_{\pm 1.99}$ & $71.90_{\pm 2.03}$ & $64.16_{\pm 2.32}$ & $55.54_{\pm 2.79}$ & $49.32_{\pm 2.66}$ & $43.90_{\pm 3.09}$ \\
    \midrule
    \multirow{4}{*}{GPRGNN} 
    & EvA & $72.01_{\pm 4.18}$ & $67.61_{\pm 4.28}$ & $55.95_{\pm 4.32}$ & $51.49_{\pm 7.51}$ & $49.18_{\pm 7.83}$ & $42.39_{\pm 9.63}$ \\
    & EvALocal & $73.01_{\pm 4.18}$ & $69.10_{\pm 3.83}$ & $62.77_{\pm 6.59}$ & $60.51_{\pm 8.10}$ & $59.72_{\pm 8.91}$ & $59.31_{\pm 9.50}$ \\
    & LRBCD & $74.50_{\pm 3.66}$ & $71.57_{\pm 4.10}$ & $65.88_{\pm 6.12}$ & $60.33_{\pm 5.70}$ & $56.75_{\pm 7.74}$ & $53.75_{\pm 8.18}$ \\
    & PRBCD & $74.37_{\pm 3.40}$ & $71.66_{\pm 3.55}$ & $64.51_{\pm 4.94}$ & $56.21_{\pm 6.46}$ & $50.26_{\pm 7.41}$ & $45.81_{\pm 8.47}$ \\
    \bottomrule
    \end{tabular}
    \label{tab:pubmed-attack}

\end{table}

%% file: tables/citeseer_models.tex
\begin{table}[ht]
    \centering
    \setlength{\tabcolsep}{4pt} 
    \caption{Classification accuracy (\%) on the \texttt{Citeseer} dataset in the inductive setting under adversarial attacks. 
    Results are reported for four GNN models subjected to two different attack methods across varying perturbation budgets $\epsilon$.
    Training, validation, and test sets are non-stratified.}
    \label{tab:citeseer-model}
    \vskip 0.5em
    \begin{tabular}{llcccccc}
    \toprule
    \multirow{2}{*}{Model} & \multirow{2}{*}{Attack} & \multicolumn{6}{c}{$\epsilon$} \\
    \cmidrule(lr){3-8}
    & & 0.01 & 0.02 & 0.05 & 0.10 & 0.15 & 0.20 \\
    \midrule
    \multirow{2}{*}{APPNP}
    & EvA & $86.90_{\pm 1.11}$ & $83.93_{\pm 0.94}$ & $74.29_{\pm 0.88}$ & $65.00_{\pm 1.15}$ & $59.76_{\pm 2.33}$ & $54.76_{\pm 1.19}$ \\
    & PRBCD & $87.26_{\pm 0.90}$ & $85.48_{\pm 1.49}$ & $81.79_{\pm 1.08}$ & $76.55_{\pm 0.68}$ & $72.44_{\pm 1.66}$ & $69.29_{\pm 1.69}$ \\
    \midrule
    \multirow{2}{*}{GAT}
    & EvA & $81.54_{\pm 1.39}$ & $76.78_{\pm 2.22}$ & $67.14_{\pm 3.65}$ & $51.19_{\pm 4.21}$ & $37.74_{\pm 4.94}$ & $27.62_{\pm 9.49}$ \\
    & PRBCD & $84.52_{\pm 2.27}$ & $82.62_{\pm 2.20}$ & $76.55_{\pm 5.59}$ & $70.00_{\pm 6.09}$ & $67.02_{\pm 4.27}$ & $63.15_{\pm 4.19}$ \\
    \midrule
    \multirow{2}{*}{GCN}
    & EvA & $86.67_{\pm 1.71}$ & $82.86_{\pm 2.12}$ & $72.74_{\pm 2.74}$ & $58.33_{\pm 3.01}$ & $49.76_{\pm 3.22}$ & $44.29_{\pm 3.33}$ \\
    & PRBCD & $87.38_{\pm 1.81}$ & $85.83_{\pm 2.43}$ & $80.95_{\pm 2.06}$ & $74.29_{\pm 4.22}$ & $69.76_{\pm 4.34}$ & $67.62_{\pm 4.96}$ \\
    \midrule
    \multirow{2}{*}{GPRGNN}
    & EvA & $87.26_{\pm 2.75}$ & $83.81_{\pm 2.50}$ & $73.45_{\pm 3.17}$ & $61.43_{\pm 4.66}$ & $55.48_{\pm 3.84}$ & $50.12_{\pm 4.86}$ \\
    & PRBCD & $88.45_{\pm 2.29}$ & $86.31_{\pm 2.45}$ & $82.02_{\pm 2.61}$ & $77.14_{\pm 2.84}$ & $73.93_{\pm 3.89}$ & $69.64_{\pm 3.47}$ \\
    \bottomrule
    \end{tabular}
\end{table}

%% file: tables/citeseer_attacks.tex
\begin{table}[ht]
    \centering
    \setlength{\tabcolsep}{4pt} 
    \caption{Classification accuracy (\%) on the \texttt{Citeseer} dataset in the inductive setting under adversarial attacks. 
    Results are reported for two GNN models subjected to four different attack methods across varying perturbation budgets $\epsilon$.
    Training, validation, and test sets are non-stratified.}
    \label{tab:citeseer-attack}
    \vskip 0.5em
    \begin{tabular}{llcccccc}
    \toprule
    \multirow{2}{*}{Model} & \multirow{2}{*}{Attack} & \multicolumn{6}{c}{$\epsilon$} \\
    \cmidrule(lr){3-8}
    & & 0.01 & 0.02 & 0.05 & 0.10 & 0.15 & 0.20 \\
    \midrule
    \multirow{4}{*}{GCN}
    & EvA & $86.67_{\pm 1.71}$ & $82.86_{\pm 2.12}$ & $72.74_{\pm 2.74}$ & $58.33_{\pm 3.01}$ & $49.76_{\pm 3.22}$ & $44.29_{\pm 3.33}$ \\
    & EvaLocal & $87.38_{\pm 1.65}$ & $83.57_{\pm 2.17}$ & $78.21_{\pm 3.17}$ & $76.43_{\pm 2.62}$ & $75.00_{\pm 3.23}$ & $74.52_{\pm 3.16}$ \\
    & LRBCD & $88.45_{\pm 2.17}$ & $86.43_{\pm 2.71}$ & $83.69_{\pm 2.48}$ & $80.12_{\pm 3.30}$ & $78.45_{\pm 3.89}$ & $75.36_{\pm 4.81}$ \\
    & PRBCD & $87.38_{\pm 1.81}$ & $85.83_{\pm 2.43}$ & $80.95_{\pm 2.06}$ & $74.29_{\pm 4.22}$ & $69.76_{\pm 4.34}$ & $67.62_{\pm 4.96}$ \\
    \midrule
    \multirow{4}{*}{GPRGNN}
    & EvA & $87.26_{\pm 2.75}$ & $83.81_{\pm 2.50}$ & $73.45_{\pm 3.17}$ & $61.43_{\pm 4.66}$ & $55.48_{\pm 3.84}$ & $50.12_{\pm 4.86}$ \\
    & EvaLocal & $87.50_{\pm 2.27}$ & $84.29_{\pm 2.04}$ & $80.48_{\pm 3.96}$ & $77.86_{\pm 4.56}$ & $76.43_{\pm 5.06}$ & $75.12_{\pm 6.57}$ \\
    & LRBCD & $89.76_{\pm 2.50}$ & $87.98_{\pm 2.48}$ & $85.12_{\pm 2.76}$ & $81.90_{\pm 2.83}$ & $79.64_{\pm 4.08}$ & $78.45_{\pm 4.92}$ \\
    & PRBCD & $88.45_{\pm 2.29}$ & $86.31_{\pm 2.45}$ & $82.02_{\pm 2.61}$ & $77.14_{\pm 2.84}$ & $73.93_{\pm 3.89}$ & $69.64_{\pm 3.47}$ \\
    \bottomrule
    \end{tabular}
\end{table}

%% file: tables/results_table.tex
\begin{table}[ht]
    \centering
    \setlength{\tabcolsep}{3pt} 
    \caption{Classification accuracy (\%) on the \texttt{CoraML} dataset in the inductive setting under adversarial attacks. 
    Results are reported for two GNN models with or without adversarial training subjected to three different attack methods across varying perturbation budgets $\epsilon$.
    Training, validation, and test sets are stratified.}
    \label{tab:coraml_inductive}
    \vskip 0.5em
    \begin{tabular}{lllcccccc}
    \toprule
    \multirow{2}{*}{Model} & \multirow{2}{*}{Adv.\ Tr.} & \multirow{2}{*}{Attack} & \multicolumn{6}{c}{$\epsilon$} \\
    \cmidrule(lr){4-9}
    & & & 0.01 & 0.02 & 0.05 & 0.10 & 0.15 & 0.20 \\
    \midrule
    \multirow{12}{*}{\rotatebox{90}{GCN}} 
    & \multirow{3}{*}{None} 
        & LRBCD & $78.51_{\pm 1.56}$ & $75.94_{\pm 1.54}$ & $71.10_{\pm 1.16}$ & $64.41_{\pm 1.65}$ & $60.14_{\pm 1.73}$ & $57.37_{\pm 1.45}$ \\
    & & PRBCD & $76.44_{\pm 1.64}$ & $73.17_{\pm 1.39}$ & $66.48_{\pm 2.13}$ & $58.51_{\pm 1.77}$ & $52.67_{\pm 2.09}$ & $47.19_{\pm 2.02}$ \\
    & & EvA & $74.80_{\pm 1.50}$ & $68.97_{\pm 1.58}$ & $52.95_{\pm 1.91}$ & $41.99_{\pm 2.06}$ & $37.65_{\pm 2.74}$ & $35.37_{\pm 2.38}$ \\
    \cmidrule(lr){2-9}
    & \multirow{3}{*}{LRBCD} 
        & LRBCD & $79.64_{\pm 1.77}$ & $77.51_{\pm 2.41}$ & $73.10_{\pm 1.54}$ & $68.19_{\pm 1.11}$ & $64.84_{\pm 1.92}$ & $62.35_{\pm 3.00}$ \\
    & & PRBCD & $78.79_{\pm 1.88}$ & $75.87_{\pm 1.41}$ & $69.75_{\pm 1.81}$ & $62.35_{\pm 2.70}$ & $56.80_{\pm 3.04}$ & $54.23_{\pm 4.71}$ \\
    & & EvA & $76.80_{\pm 1.29}$ & $71.10_{\pm 1.64}$ & $56.30_{\pm 1.66}$ & $48.40_{\pm 2.91}$ & $43.06_{\pm 2.43}$ & $40.85_{\pm 2.70}$ \\
    \cmidrule(lr){2-9}
    & \multirow{3}{*}{PRBCD} 
        & LRBCD & $80.71_{\pm 1.16}$ & $77.86_{\pm 0.81}$ & $73.81_{\pm 0.54}$ & $69.40_{\pm 0.91}$ & $66.48_{\pm 1.08}$ & $63.77_{\pm 1.73}$ \\
    & & PRBCD & $78.93_{\pm 1.27}$ & $76.30_{\pm 1.27}$ & $70.25_{\pm 1.74}$ & $64.06_{\pm 1.83}$ & $59.50_{\pm 2.84}$ & $56.58_{\pm 4.53}$ \\
    & & EvA & $76.80_{\pm 0.77}$ & $71.53_{\pm 1.65}$ & $57.44_{\pm 2.13}$ & $49.11_{\pm 3.05}$ & $44.70_{\pm 2.96}$ & $41.92_{\pm 3.32}$ \\
    \cmidrule(lr){2-9}
    & \multirow{3}{*}{EvA} 
        & LRBCD & $80.85_{\pm 1.36}$ & $78.58_{\pm 0.99}$ & $74.66_{\pm 1.11}$ & $69.89_{\pm 0.93}$ & $66.98_{\pm 0.92}$ & $65.05_{\pm 1.08}$ \\
    & & PRBCD & $79.79_{\pm 1.80}$ & $76.51_{\pm 1.31}$ & $71.25_{\pm 1.54}$ & $64.34_{\pm 1.97}$ & $60.43_{\pm 1.32}$ & $58.22_{\pm 2.26}$ \\
    & & EvA & $77.22_{\pm 1.87}$ & $71.96_{\pm 2.38}$ & $57.94_{\pm 3.08}$ & $50.04_{\pm 3.29}$ & $44.91_{\pm 3.44}$ & $42.63_{\pm 2.33}$ \\
    \midrule
    \multirow{12}{*}{\rotatebox{90}{GPRGNN}} 
    & \multirow{3}{*}{None} 
        & LRBCD & $77.51_{\pm 2.81}$ & $74.80_{\pm 3.08}$ & $68.83_{\pm 4.20}$ & $62.56_{\pm 4.69}$ & $59.07_{\pm 5.98}$ & $55.66_{\pm 6.99}$ \\
    & & PRBCD & $74.95_{\pm 3.08}$ & $71.67_{\pm 2.76}$ & $64.84_{\pm 4.18}$ & $57.94_{\pm 4.55}$ & $53.24_{\pm 5.20}$ & $48.68_{\pm 6.52}$ \\
    & & EvA & $72.53_{\pm 4.11}$ & $66.83_{\pm 4.54}$ & $51.53_{\pm 5.57}$ & $42.21_{\pm 8.52}$ & $37.01_{\pm 9.84}$ & $34.52_{\pm 9.83}$ \\
    \cmidrule(lr){2-9}
    & \multirow{3}{*}{LRBCD} 
        & LRBCD & $81.57_{\pm 2.58}$ & $79.72_{\pm 2.22}$ & $75.59_{\pm 2.31}$ & $71.32_{\pm 2.20}$ & $68.97_{\pm 2.10}$ & $66.69_{\pm 2.25}$ \\
    & & PRBCD & $80.71_{\pm 2.61}$ & $78.51_{\pm 2.29}$ & $72.88_{\pm 2.38}$ & $66.90_{\pm 1.95}$ & $61.78_{\pm 1.99}$ & $57.51_{\pm 3.72}$ \\
    & & EvA & $78.79_{\pm 2.69}$ & $72.95_{\pm 2.67}$ & $63.42_{\pm 3.15}$ & $56.58_{\pm 4.68}$ & $52.88_{\pm 5.61}$ & $49.96_{\pm 5.75}$ \\
    \cmidrule(lr){2-9}
    & \multirow{3}{*}{PRBCD} 
        & LRBCD & $80.43_{\pm 2.01}$ & $78.01_{\pm 1.91}$ & $73.74_{\pm 1.66}$ & $69.96_{\pm 2.14}$ & $67.19_{\pm 2.51}$ & $64.84_{\pm 3.20}$ \\
    & & PRBCD & $80.21_{\pm 2.43}$ & $77.30_{\pm 2.63}$ & $71.53_{\pm 2.67}$ & $65.12_{\pm 3.21}$ & $60.07_{\pm 4.10}$ & $55.37_{\pm 3.85}$ \\
    & & EvA & $78.79_{\pm 2.45}$ & $73.10_{\pm 2.54}$ & $62.85_{\pm 4.93}$ & $56.94_{\pm 6.64}$ & $53.74_{\pm 7.65}$ & $51.60_{\pm 8.10}$ \\
    \cmidrule(lr){2-9}
    & \multirow{3}{*}{EvA} 
        & LRBCD & $79.64_{\pm 0.89}$ & $76.44_{\pm 0.68}$ & $72.95_{\pm 1.04}$ & $69.04_{\pm 1.26}$ & $67.05_{\pm 1.46}$ & $65.48_{\pm 1.88}$ \\
    & & PRBCD & $78.51_{\pm 0.60}$ & $75.87_{\pm 1.32}$ & $70.32_{\pm 0.89}$ & $64.91_{\pm 1.14}$ & $59.57_{\pm 1.75}$ & $56.16_{\pm 1.62}$ \\
    & & EvA & $76.51_{\pm 0.44}$ & $70.96_{\pm 0.41}$ & $60.85_{\pm 3.07}$ & $54.73_{\pm 3.99}$ & $50.25_{\pm 5.57}$ & $48.83_{\pm 5.95}$ \\
    \bottomrule
        \end{tabular}
\end{table}

%% file: figures/dc_scaling.pgf
\begingroup%
\makeatletter%
\begin{pgfpicture}%
\pgfpathrectangle{\pgfpointorigin}{\pgfqpoint{2.275000in}{1.700000in}}%
\pgfusepath{use as bounding box, clip}%
\begin{pgfscope}%
\pgfsetbuttcap%
\pgfsetmiterjoin%
\definecolor{currentfill}{rgb}{1.000000,1.000000,1.000000}%
\pgfsetfillcolor{currentfill}%
\pgfsetlinewidth{0.000000pt}%
\definecolor{currentstroke}{rgb}{1.000000,1.000000,1.000000}%
\pgfsetstrokecolor{currentstroke}%
\pgfsetstrokeopacity{0.000000}%
\pgfsetdash{}{0pt}%
\pgfpathmoveto{\pgfqpoint{0.000000in}{0.000000in}}%
\pgfpathlineto{\pgfqpoint{2.275000in}{0.000000in}}%
\pgfpathlineto{\pgfqpoint{2.275000in}{1.700000in}}%
\pgfpathlineto{\pgfqpoint{0.000000in}{1.700000in}}%
\pgfpathlineto{\pgfqpoint{0.000000in}{0.000000in}}%
\pgfpathclose%
\pgfusepath{fill}%
\end{pgfscope}%
\begin{pgfscope}%
\pgfsetbuttcap%
\pgfsetmiterjoin%
\definecolor{currentfill}{rgb}{1.000000,1.000000,1.000000}%
\pgfsetfillcolor{currentfill}%
\pgfsetlinewidth{0.000000pt}%
\definecolor{currentstroke}{rgb}{0.000000,0.000000,0.000000}%
\pgfsetstrokecolor{currentstroke}%
\pgfsetstrokeopacity{0.000000}%
\pgfsetdash{}{0pt}%
\pgfpathmoveto{\pgfqpoint{0.681786in}{0.358869in}}%
\pgfpathlineto{\pgfqpoint{2.118501in}{0.358869in}}%
\pgfpathlineto{\pgfqpoint{2.118501in}{1.567520in}}%
\pgfpathlineto{\pgfqpoint{0.681786in}{1.567520in}}%
\pgfpathlineto{\pgfqpoint{0.681786in}{0.358869in}}%
\pgfpathclose%
\pgfusepath{fill}%
\end{pgfscope}%
\begin{pgfscope}%
\pgfpathrectangle{\pgfqpoint{0.681786in}{0.358869in}}{\pgfqpoint{1.436716in}{1.208651in}}%
\pgfusepath{clip}%
\pgfsetroundcap%
\pgfsetroundjoin%
\pgfsetlinewidth{0.803000pt}%
\definecolor{currentstroke}{rgb}{0.800000,0.800000,0.800000}%
\pgfsetstrokecolor{currentstroke}%
\pgfsetdash{}{0pt}%
\pgfpathmoveto{\pgfqpoint{0.747091in}{0.358869in}}%
\pgfpathlineto{\pgfqpoint{0.747091in}{1.567520in}}%
\pgfusepath{stroke}%
\end{pgfscope}%
\begin{pgfscope}%
\definecolor{textcolor}{rgb}{0.150000,0.150000,0.150000}%
\pgfsetstrokecolor{textcolor}%
\pgfsetfillcolor{textcolor}%
\pgftext[x=0.747091in,y=0.243591in,,top]{\color{textcolor}{\rmfamily\fontsize{8.096000}{9.715200}\selectfont\catcode`\^=\active\def^{\ifmmode\sp\else\^{}\fi}\catcode`\%=\active\def
\end{pgfscope}%
\begin{pgfscope}%
\pgfpathrectangle{\pgfqpoint{0.681786in}{0.358869in}}{\pgfqpoint{1.436716in}{1.208651in}}%
\pgfusepath{clip}%
\pgfsetroundcap%
\pgfsetroundjoin%
\pgfsetlinewidth{0.803000pt}%
\definecolor{currentstroke}{rgb}{0.800000,0.800000,0.800000}%
\pgfsetstrokecolor{currentstroke}%
\pgfsetdash{}{0pt}%
\pgfpathmoveto{\pgfqpoint{1.115008in}{0.358869in}}%
\pgfpathlineto{\pgfqpoint{1.115008in}{1.567520in}}%
\pgfusepath{stroke}%
\end{pgfscope}%
\begin{pgfscope}%
\definecolor{textcolor}{rgb}{0.150000,0.150000,0.150000}%
\pgfsetstrokecolor{textcolor}%
\pgfsetfillcolor{textcolor}%
\pgftext[x=1.115008in,y=0.243591in,,top]{\color{textcolor}{\rmfamily\fontsize{8.096000}{9.715200}\selectfont\catcode`\^=\active\def^{\ifmmode\sp\else\^{}\fi}\catcode`\%=\active\def
\end{pgfscope}%
\begin{pgfscope}%
\pgfpathrectangle{\pgfqpoint{0.681786in}{0.358869in}}{\pgfqpoint{1.436716in}{1.208651in}}%
\pgfusepath{clip}%
\pgfsetroundcap%
\pgfsetroundjoin%
\pgfsetlinewidth{0.803000pt}%
\definecolor{currentstroke}{rgb}{0.800000,0.800000,0.800000}%
\pgfsetstrokecolor{currentstroke}%
\pgfsetdash{}{0pt}%
\pgfpathmoveto{\pgfqpoint{1.574904in}{0.358869in}}%
\pgfpathlineto{\pgfqpoint{1.574904in}{1.567520in}}%
\pgfusepath{stroke}%
\end{pgfscope}%
\begin{pgfscope}%
\definecolor{textcolor}{rgb}{0.150000,0.150000,0.150000}%
\pgfsetstrokecolor{textcolor}%
\pgfsetfillcolor{textcolor}%
\pgftext[x=1.574904in,y=0.243591in,,top]{\color{textcolor}{\rmfamily\fontsize{8.096000}{9.715200}\selectfont\catcode`\^=\active\def^{\ifmmode\sp\else\^{}\fi}\catcode`\%=\active\def
\end{pgfscope}%
\begin{pgfscope}%
\pgfpathrectangle{\pgfqpoint{0.681786in}{0.358869in}}{\pgfqpoint{1.436716in}{1.208651in}}%
\pgfusepath{clip}%
\pgfsetroundcap%
\pgfsetroundjoin%
\pgfsetlinewidth{0.803000pt}%
\definecolor{currentstroke}{rgb}{0.800000,0.800000,0.800000}%
\pgfsetstrokecolor{currentstroke}%
\pgfsetdash{}{0pt}%
\pgfpathmoveto{\pgfqpoint{2.034800in}{0.358869in}}%
\pgfpathlineto{\pgfqpoint{2.034800in}{1.567520in}}%
\pgfusepath{stroke}%
\end{pgfscope}%
\begin{pgfscope}%
\definecolor{textcolor}{rgb}{0.150000,0.150000,0.150000}%
\pgfsetstrokecolor{textcolor}%
\pgfsetfillcolor{textcolor}%
\pgftext[x=2.034800in,y=0.243591in,,top]{\color{textcolor}{\rmfamily\fontsize{8.096000}{9.715200}\selectfont\catcode`\^=\active\def^{\ifmmode\sp\else\^{}\fi}\catcode`\%=\active\def
\end{pgfscope}%
\begin{pgfscope}%
\pgfpathrectangle{\pgfqpoint{0.681786in}{0.358869in}}{\pgfqpoint{1.436716in}{1.208651in}}%
\pgfusepath{clip}%
\pgfsetroundcap%
\pgfsetroundjoin%
\pgfsetlinewidth{0.803000pt}%
\definecolor{currentstroke}{rgb}{0.800000,0.800000,0.800000}%
\pgfsetstrokecolor{currentstroke}%
\pgfsetdash{}{0pt}%
\pgfpathmoveto{\pgfqpoint{0.681786in}{0.376592in}}%
\pgfpathlineto{\pgfqpoint{2.118501in}{0.376592in}}%
\pgfusepath{stroke}%
\end{pgfscope}%
\begin{pgfscope}%
\definecolor{textcolor}{rgb}{0.150000,0.150000,0.150000}%
\pgfsetstrokecolor{textcolor}%
\pgfsetfillcolor{textcolor}%
\pgftext[x=0.316143in, y=0.333876in, left, base]{\color{textcolor}{\rmfamily\fontsize{8.096000}{9.715200}\selectfont\catcode`\^=\active\def^{\ifmmode\sp\else\^{}\fi}\catcode`\%=\active\def
\end{pgfscope}%
\begin{pgfscope}%
\pgfpathrectangle{\pgfqpoint{0.681786in}{0.358869in}}{\pgfqpoint{1.436716in}{1.208651in}}%
\pgfusepath{clip}%
\pgfsetroundcap%
\pgfsetroundjoin%
\pgfsetlinewidth{0.803000pt}%
\definecolor{currentstroke}{rgb}{0.800000,0.800000,0.800000}%
\pgfsetstrokecolor{currentstroke}%
\pgfsetdash{}{0pt}%
\pgfpathmoveto{\pgfqpoint{0.681786in}{0.647100in}}%
\pgfpathlineto{\pgfqpoint{2.118501in}{0.647100in}}%
\pgfusepath{stroke}%
\end{pgfscope}%
\begin{pgfscope}%
\definecolor{textcolor}{rgb}{0.150000,0.150000,0.150000}%
\pgfsetstrokecolor{textcolor}%
\pgfsetfillcolor{textcolor}%
\pgftext[x=0.316143in, y=0.604385in, left, base]{\color{textcolor}{\rmfamily\fontsize{8.096000}{9.715200}\selectfont\catcode`\^=\active\def^{\ifmmode\sp\else\^{}\fi}\catcode`\%=\active\def
\end{pgfscope}%
\begin{pgfscope}%
\pgfpathrectangle{\pgfqpoint{0.681786in}{0.358869in}}{\pgfqpoint{1.436716in}{1.208651in}}%
\pgfusepath{clip}%
\pgfsetroundcap%
\pgfsetroundjoin%
\pgfsetlinewidth{0.803000pt}%
\definecolor{currentstroke}{rgb}{0.800000,0.800000,0.800000}%
\pgfsetstrokecolor{currentstroke}%
\pgfsetdash{}{0pt}%
\pgfpathmoveto{\pgfqpoint{0.681786in}{0.917609in}}%
\pgfpathlineto{\pgfqpoint{2.118501in}{0.917609in}}%
\pgfusepath{stroke}%
\end{pgfscope}%
\begin{pgfscope}%
\definecolor{textcolor}{rgb}{0.150000,0.150000,0.150000}%
\pgfsetstrokecolor{textcolor}%
\pgfsetfillcolor{textcolor}%
\pgftext[x=0.316143in, y=0.874893in, left, base]{\color{textcolor}{\rmfamily\fontsize{8.096000}{9.715200}\selectfont\catcode`\^=\active\def^{\ifmmode\sp\else\^{}\fi}\catcode`\%=\active\def
\end{pgfscope}%
\begin{pgfscope}%
\pgfpathrectangle{\pgfqpoint{0.681786in}{0.358869in}}{\pgfqpoint{1.436716in}{1.208651in}}%
\pgfusepath{clip}%
\pgfsetroundcap%
\pgfsetroundjoin%
\pgfsetlinewidth{0.803000pt}%
\definecolor{currentstroke}{rgb}{0.800000,0.800000,0.800000}%
\pgfsetstrokecolor{currentstroke}%
\pgfsetdash{}{0pt}%
\pgfpathmoveto{\pgfqpoint{0.681786in}{1.188117in}}%
\pgfpathlineto{\pgfqpoint{2.118501in}{1.188117in}}%
\pgfusepath{stroke}%
\end{pgfscope}%
\begin{pgfscope}%
\definecolor{textcolor}{rgb}{0.150000,0.150000,0.150000}%
\pgfsetstrokecolor{textcolor}%
\pgfsetfillcolor{textcolor}%
\pgftext[x=0.316143in, y=1.145401in, left, base]{\color{textcolor}{\rmfamily\fontsize{8.096000}{9.715200}\selectfont\catcode`\^=\active\def^{\ifmmode\sp\else\^{}\fi}\catcode`\%=\active\def
\end{pgfscope}%
\begin{pgfscope}%
\pgfpathrectangle{\pgfqpoint{0.681786in}{0.358869in}}{\pgfqpoint{1.436716in}{1.208651in}}%
\pgfusepath{clip}%
\pgfsetroundcap%
\pgfsetroundjoin%
\pgfsetlinewidth{0.803000pt}%
\definecolor{currentstroke}{rgb}{0.800000,0.800000,0.800000}%
\pgfsetstrokecolor{currentstroke}%
\pgfsetdash{}{0pt}%
\pgfpathmoveto{\pgfqpoint{0.681786in}{1.458625in}}%
\pgfpathlineto{\pgfqpoint{2.118501in}{1.458625in}}%
\pgfusepath{stroke}%
\end{pgfscope}%
\begin{pgfscope}%
\definecolor{textcolor}{rgb}{0.150000,0.150000,0.150000}%
\pgfsetstrokecolor{textcolor}%
\pgfsetfillcolor{textcolor}%
\pgftext[x=0.316143in, y=1.415910in, left, base]{\color{textcolor}{\rmfamily\fontsize{8.096000}{9.715200}\selectfont\catcode`\^=\active\def^{\ifmmode\sp\else\^{}\fi}\catcode`\%=\active\def
\end{pgfscope}%
\begin{pgfscope}%
\definecolor{textcolor}{rgb}{0.150000,0.150000,0.150000}%
\pgfsetstrokecolor{textcolor}%
\pgfsetfillcolor{textcolor}%
\pgftext[x=0.260588in,y=0.963194in,,bottom,rotate=90.000000]{\color{textcolor}{\rmfamily\fontsize{8.832000}{10.598400}\selectfont\catcode`\^=\active\def^{\ifmmode\sp\else\^{}\fi}\catcode`\%=\active\def
\end{pgfscope}%
\begin{pgfscope}%
\pgfpathrectangle{\pgfqpoint{0.681786in}{0.358869in}}{\pgfqpoint{1.436716in}{1.208651in}}%
\pgfusepath{clip}%
\pgfsetbuttcap%
\pgfsetroundjoin%
\pgfsetlinewidth{1.003750pt}%
\definecolor{currentstroke}{rgb}{0.882353,0.341176,0.349020}%
\pgfsetstrokecolor{currentstroke}%
\pgfsetdash{}{0pt}%
\pgfpathmoveto{\pgfqpoint{0.765487in}{1.482823in}}%
\pgfpathlineto{\pgfqpoint{0.765487in}{1.512581in}}%
\pgfusepath{stroke}%
\end{pgfscope}%
\begin{pgfscope}%
\pgfpathrectangle{\pgfqpoint{0.681786in}{0.358869in}}{\pgfqpoint{1.436716in}{1.208651in}}%
\pgfusepath{clip}%
\pgfsetbuttcap%
\pgfsetroundjoin%
\pgfsetlinewidth{1.003750pt}%
\definecolor{currentstroke}{rgb}{0.882353,0.341176,0.349020}%
\pgfsetstrokecolor{currentstroke}%
\pgfsetdash{}{0pt}%
\pgfpathmoveto{\pgfqpoint{0.857466in}{1.336313in}}%
\pgfpathlineto{\pgfqpoint{0.857466in}{1.354240in}}%
\pgfusepath{stroke}%
\end{pgfscope}%
\begin{pgfscope}%
\pgfpathrectangle{\pgfqpoint{0.681786in}{0.358869in}}{\pgfqpoint{1.436716in}{1.208651in}}%
\pgfusepath{clip}%
\pgfsetbuttcap%
\pgfsetroundjoin%
\pgfsetlinewidth{1.003750pt}%
\definecolor{currentstroke}{rgb}{0.882353,0.341176,0.349020}%
\pgfsetstrokecolor{currentstroke}%
\pgfsetdash{}{0pt}%
\pgfpathmoveto{\pgfqpoint{1.133404in}{1.039317in}}%
\pgfpathlineto{\pgfqpoint{1.133404in}{1.065282in}}%
\pgfusepath{stroke}%
\end{pgfscope}%
\begin{pgfscope}%
\pgfpathrectangle{\pgfqpoint{0.681786in}{0.358869in}}{\pgfqpoint{1.436716in}{1.208651in}}%
\pgfusepath{clip}%
\pgfsetbuttcap%
\pgfsetroundjoin%
\pgfsetlinewidth{1.003750pt}%
\definecolor{currentstroke}{rgb}{0.882353,0.341176,0.349020}%
\pgfsetstrokecolor{currentstroke}%
\pgfsetdash{}{0pt}%
\pgfpathmoveto{\pgfqpoint{1.593300in}{0.730382in}}%
\pgfpathlineto{\pgfqpoint{1.593300in}{0.758729in}}%
\pgfusepath{stroke}%
\end{pgfscope}%
\begin{pgfscope}%
\pgfpathrectangle{\pgfqpoint{0.681786in}{0.358869in}}{\pgfqpoint{1.436716in}{1.208651in}}%
\pgfusepath{clip}%
\pgfsetbuttcap%
\pgfsetroundjoin%
\pgfsetlinewidth{1.003750pt}%
\definecolor{currentstroke}{rgb}{0.882353,0.341176,0.349020}%
\pgfsetstrokecolor{currentstroke}%
\pgfsetdash{}{0pt}%
\pgfpathmoveto{\pgfqpoint{2.053196in}{0.532172in}}%
\pgfpathlineto{\pgfqpoint{2.053196in}{0.557323in}}%
\pgfusepath{stroke}%
\end{pgfscope}%
\begin{pgfscope}%
\pgfpathrectangle{\pgfqpoint{0.681786in}{0.358869in}}{\pgfqpoint{1.436716in}{1.208651in}}%
\pgfusepath{clip}%
\pgfsetbuttcap%
\pgfsetroundjoin%
\definecolor{currentfill}{rgb}{0.882353,0.341176,0.349020}%
\pgfsetfillcolor{currentfill}%
\pgfsetlinewidth{1.003750pt}%
\definecolor{currentstroke}{rgb}{0.882353,0.341176,0.349020}%
\pgfsetstrokecolor{currentstroke}%
\pgfsetdash{}{0pt}%
\pgfsys@defobject{currentmarker}{\pgfqpoint{-0.027778in}{-0.000000in}}{\pgfqpoint{0.027778in}{0.000000in}}{%
\pgfpathmoveto{\pgfqpoint{0.027778in}{-0.000000in}}%
\pgfpathlineto{\pgfqpoint{-0.027778in}{0.000000in}}%
\pgfusepath{stroke,fill}%
}%
\begin{pgfscope}%
\pgfsys@transformshift{0.765487in}{1.482823in}%
\pgfsys@useobject{currentmarker}{}%
\end{pgfscope}%
\begin{pgfscope}%
\pgfsys@transformshift{0.857466in}{1.336313in}%
\pgfsys@useobject{currentmarker}{}%
\end{pgfscope}%
\begin{pgfscope}%
\pgfsys@transformshift{1.133404in}{1.039317in}%
\pgfsys@useobject{currentmarker}{}%
\end{pgfscope}%
\begin{pgfscope}%
\pgfsys@transformshift{1.593300in}{0.730382in}%
\pgfsys@useobject{currentmarker}{}%
\end{pgfscope}%
\begin{pgfscope}%
\pgfsys@transformshift{2.053196in}{0.532172in}%
\pgfsys@useobject{currentmarker}{}%
\end{pgfscope}%
\end{pgfscope}%
\begin{pgfscope}%
\pgfpathrectangle{\pgfqpoint{0.681786in}{0.358869in}}{\pgfqpoint{1.436716in}{1.208651in}}%
\pgfusepath{clip}%
\pgfsetbuttcap%
\pgfsetroundjoin%
\definecolor{currentfill}{rgb}{0.882353,0.341176,0.349020}%
\pgfsetfillcolor{currentfill}%
\pgfsetlinewidth{1.003750pt}%
\definecolor{currentstroke}{rgb}{0.882353,0.341176,0.349020}%
\pgfsetstrokecolor{currentstroke}%
\pgfsetdash{}{0pt}%
\pgfsys@defobject{currentmarker}{\pgfqpoint{-0.027778in}{-0.000000in}}{\pgfqpoint{0.027778in}{0.000000in}}{%
\pgfpathmoveto{\pgfqpoint{0.027778in}{-0.000000in}}%
\pgfpathlineto{\pgfqpoint{-0.027778in}{0.000000in}}%
\pgfusepath{stroke,fill}%
}%
\begin{pgfscope}%
\pgfsys@transformshift{0.765487in}{1.512581in}%
\pgfsys@useobject{currentmarker}{}%
\end{pgfscope}%
\begin{pgfscope}%
\pgfsys@transformshift{0.857466in}{1.354240in}%
\pgfsys@useobject{currentmarker}{}%
\end{pgfscope}%
\begin{pgfscope}%
\pgfsys@transformshift{1.133404in}{1.065282in}%
\pgfsys@useobject{currentmarker}{}%
\end{pgfscope}%
\begin{pgfscope}%
\pgfsys@transformshift{1.593300in}{0.758729in}%
\pgfsys@useobject{currentmarker}{}%
\end{pgfscope}%
\begin{pgfscope}%
\pgfsys@transformshift{2.053196in}{0.557323in}%
\pgfsys@useobject{currentmarker}{}%
\end{pgfscope}%
\end{pgfscope}%
\begin{pgfscope}%
\pgfpathrectangle{\pgfqpoint{0.681786in}{0.358869in}}{\pgfqpoint{1.436716in}{1.208651in}}%
\pgfusepath{clip}%
\pgfsetbuttcap%
\pgfsetroundjoin%
\pgfsetlinewidth{1.003750pt}%
\definecolor{currentstroke}{rgb}{0.882353,0.341176,0.349020}%
\pgfsetstrokecolor{currentstroke}%
\pgfsetdash{}{0pt}%
\pgfpathmoveto{\pgfqpoint{0.765487in}{1.396184in}}%
\pgfpathlineto{\pgfqpoint{0.765487in}{1.425942in}}%
\pgfusepath{stroke}%
\end{pgfscope}%
\begin{pgfscope}%
\pgfpathrectangle{\pgfqpoint{0.681786in}{0.358869in}}{\pgfqpoint{1.436716in}{1.208651in}}%
\pgfusepath{clip}%
\pgfsetbuttcap%
\pgfsetroundjoin%
\pgfsetlinewidth{1.003750pt}%
\definecolor{currentstroke}{rgb}{0.882353,0.341176,0.349020}%
\pgfsetstrokecolor{currentstroke}%
\pgfsetdash{}{0pt}%
\pgfpathmoveto{\pgfqpoint{0.857466in}{1.215278in}}%
\pgfpathlineto{\pgfqpoint{0.857466in}{1.233206in}}%
\pgfusepath{stroke}%
\end{pgfscope}%
\begin{pgfscope}%
\pgfpathrectangle{\pgfqpoint{0.681786in}{0.358869in}}{\pgfqpoint{1.436716in}{1.208651in}}%
\pgfusepath{clip}%
\pgfsetbuttcap%
\pgfsetroundjoin%
\pgfsetlinewidth{1.003750pt}%
\definecolor{currentstroke}{rgb}{0.882353,0.341176,0.349020}%
\pgfsetstrokecolor{currentstroke}%
\pgfsetdash{}{0pt}%
\pgfpathmoveto{\pgfqpoint{1.133404in}{0.943773in}}%
\pgfpathlineto{\pgfqpoint{1.133404in}{0.969738in}}%
\pgfusepath{stroke}%
\end{pgfscope}%
\begin{pgfscope}%
\pgfpathrectangle{\pgfqpoint{0.681786in}{0.358869in}}{\pgfqpoint{1.436716in}{1.208651in}}%
\pgfusepath{clip}%
\pgfsetbuttcap%
\pgfsetroundjoin%
\pgfsetlinewidth{1.003750pt}%
\definecolor{currentstroke}{rgb}{0.882353,0.341176,0.349020}%
\pgfsetstrokecolor{currentstroke}%
\pgfsetdash{}{0pt}%
\pgfpathmoveto{\pgfqpoint{1.593300in}{0.689715in}}%
\pgfpathlineto{\pgfqpoint{1.593300in}{0.718062in}}%
\pgfusepath{stroke}%
\end{pgfscope}%
\begin{pgfscope}%
\pgfpathrectangle{\pgfqpoint{0.681786in}{0.358869in}}{\pgfqpoint{1.436716in}{1.208651in}}%
\pgfusepath{clip}%
\pgfsetbuttcap%
\pgfsetroundjoin%
\pgfsetlinewidth{1.003750pt}%
\definecolor{currentstroke}{rgb}{0.882353,0.341176,0.349020}%
\pgfsetstrokecolor{currentstroke}%
\pgfsetdash{}{0pt}%
\pgfpathmoveto{\pgfqpoint{2.053196in}{0.513620in}}%
\pgfpathlineto{\pgfqpoint{2.053196in}{0.538771in}}%
\pgfusepath{stroke}%
\end{pgfscope}%
\begin{pgfscope}%
\pgfpathrectangle{\pgfqpoint{0.681786in}{0.358869in}}{\pgfqpoint{1.436716in}{1.208651in}}%
\pgfusepath{clip}%
\pgfsetbuttcap%
\pgfsetroundjoin%
\definecolor{currentfill}{rgb}{0.882353,0.341176,0.349020}%
\pgfsetfillcolor{currentfill}%
\pgfsetlinewidth{1.003750pt}%
\definecolor{currentstroke}{rgb}{0.882353,0.341176,0.349020}%
\pgfsetstrokecolor{currentstroke}%
\pgfsetdash{}{0pt}%
\pgfsys@defobject{currentmarker}{\pgfqpoint{-0.027778in}{-0.000000in}}{\pgfqpoint{0.027778in}{0.000000in}}{%
\pgfpathmoveto{\pgfqpoint{0.027778in}{-0.000000in}}%
\pgfpathlineto{\pgfqpoint{-0.027778in}{0.000000in}}%
\pgfusepath{stroke,fill}%
}%
\begin{pgfscope}%
\pgfsys@transformshift{0.765487in}{1.396184in}%
\pgfsys@useobject{currentmarker}{}%
\end{pgfscope}%
\begin{pgfscope}%
\pgfsys@transformshift{0.857466in}{1.215278in}%
\pgfsys@useobject{currentmarker}{}%
\end{pgfscope}%
\begin{pgfscope}%
\pgfsys@transformshift{1.133404in}{0.943773in}%
\pgfsys@useobject{currentmarker}{}%
\end{pgfscope}%
\begin{pgfscope}%
\pgfsys@transformshift{1.593300in}{0.689715in}%
\pgfsys@useobject{currentmarker}{}%
\end{pgfscope}%
\begin{pgfscope}%
\pgfsys@transformshift{2.053196in}{0.513620in}%
\pgfsys@useobject{currentmarker}{}%
\end{pgfscope}%
\end{pgfscope}%
\begin{pgfscope}%
\pgfpathrectangle{\pgfqpoint{0.681786in}{0.358869in}}{\pgfqpoint{1.436716in}{1.208651in}}%
\pgfusepath{clip}%
\pgfsetbuttcap%
\pgfsetroundjoin%
\definecolor{currentfill}{rgb}{0.882353,0.341176,0.349020}%
\pgfsetfillcolor{currentfill}%
\pgfsetlinewidth{1.003750pt}%
\definecolor{currentstroke}{rgb}{0.882353,0.341176,0.349020}%
\pgfsetstrokecolor{currentstroke}%
\pgfsetdash{}{0pt}%
\pgfsys@defobject{currentmarker}{\pgfqpoint{-0.027778in}{-0.000000in}}{\pgfqpoint{0.027778in}{0.000000in}}{%
\pgfpathmoveto{\pgfqpoint{0.027778in}{-0.000000in}}%
\pgfpathlineto{\pgfqpoint{-0.027778in}{0.000000in}}%
\pgfusepath{stroke,fill}%
}%
\begin{pgfscope}%
\pgfsys@transformshift{0.765487in}{1.425942in}%
\pgfsys@useobject{currentmarker}{}%
\end{pgfscope}%
\begin{pgfscope}%
\pgfsys@transformshift{0.857466in}{1.233206in}%
\pgfsys@useobject{currentmarker}{}%
\end{pgfscope}%
\begin{pgfscope}%
\pgfsys@transformshift{1.133404in}{0.969738in}%
\pgfsys@useobject{currentmarker}{}%
\end{pgfscope}%
\begin{pgfscope}%
\pgfsys@transformshift{1.593300in}{0.718062in}%
\pgfsys@useobject{currentmarker}{}%
\end{pgfscope}%
\begin{pgfscope}%
\pgfsys@transformshift{2.053196in}{0.538771in}%
\pgfsys@useobject{currentmarker}{}%
\end{pgfscope}%
\end{pgfscope}%
\begin{pgfscope}%
\pgfpathrectangle{\pgfqpoint{0.681786in}{0.358869in}}{\pgfqpoint{1.436716in}{1.208651in}}%
\pgfusepath{clip}%
\pgfsetbuttcap%
\pgfsetroundjoin%
\pgfsetlinewidth{1.003750pt}%
\definecolor{currentstroke}{rgb}{0.882353,0.341176,0.349020}%
\pgfsetstrokecolor{currentstroke}%
\pgfsetdash{}{0pt}%
\pgfpathmoveto{\pgfqpoint{0.747091in}{1.339505in}}%
\pgfpathlineto{\pgfqpoint{0.747091in}{1.367152in}}%
\pgfusepath{stroke}%
\end{pgfscope}%
\begin{pgfscope}%
\pgfpathrectangle{\pgfqpoint{0.681786in}{0.358869in}}{\pgfqpoint{1.436716in}{1.208651in}}%
\pgfusepath{clip}%
\pgfsetbuttcap%
\pgfsetroundjoin%
\pgfsetlinewidth{1.003750pt}%
\definecolor{currentstroke}{rgb}{0.882353,0.341176,0.349020}%
\pgfsetstrokecolor{currentstroke}%
\pgfsetdash{}{0pt}%
\pgfpathmoveto{\pgfqpoint{0.839070in}{1.093538in}}%
\pgfpathlineto{\pgfqpoint{0.839070in}{1.129944in}}%
\pgfusepath{stroke}%
\end{pgfscope}%
\begin{pgfscope}%
\pgfpathrectangle{\pgfqpoint{0.681786in}{0.358869in}}{\pgfqpoint{1.436716in}{1.208651in}}%
\pgfusepath{clip}%
\pgfsetbuttcap%
\pgfsetroundjoin%
\pgfsetlinewidth{1.003750pt}%
\definecolor{currentstroke}{rgb}{0.882353,0.341176,0.349020}%
\pgfsetstrokecolor{currentstroke}%
\pgfsetdash{}{0pt}%
\pgfpathmoveto{\pgfqpoint{1.115008in}{0.764217in}}%
\pgfpathlineto{\pgfqpoint{1.115008in}{0.793166in}}%
\pgfusepath{stroke}%
\end{pgfscope}%
\begin{pgfscope}%
\pgfpathrectangle{\pgfqpoint{0.681786in}{0.358869in}}{\pgfqpoint{1.436716in}{1.208651in}}%
\pgfusepath{clip}%
\pgfsetbuttcap%
\pgfsetroundjoin%
\pgfsetlinewidth{1.003750pt}%
\definecolor{currentstroke}{rgb}{0.882353,0.341176,0.349020}%
\pgfsetstrokecolor{currentstroke}%
\pgfsetdash{}{0pt}%
\pgfpathmoveto{\pgfqpoint{1.574904in}{0.531776in}}%
\pgfpathlineto{\pgfqpoint{1.574904in}{0.574120in}}%
\pgfusepath{stroke}%
\end{pgfscope}%
\begin{pgfscope}%
\pgfpathrectangle{\pgfqpoint{0.681786in}{0.358869in}}{\pgfqpoint{1.436716in}{1.208651in}}%
\pgfusepath{clip}%
\pgfsetbuttcap%
\pgfsetroundjoin%
\pgfsetlinewidth{1.003750pt}%
\definecolor{currentstroke}{rgb}{0.882353,0.341176,0.349020}%
\pgfsetstrokecolor{currentstroke}%
\pgfsetdash{}{0pt}%
\pgfpathmoveto{\pgfqpoint{2.034800in}{0.413808in}}%
\pgfpathlineto{\pgfqpoint{2.034800in}{0.436661in}}%
\pgfusepath{stroke}%
\end{pgfscope}%
\begin{pgfscope}%
\pgfpathrectangle{\pgfqpoint{0.681786in}{0.358869in}}{\pgfqpoint{1.436716in}{1.208651in}}%
\pgfusepath{clip}%
\pgfsetbuttcap%
\pgfsetroundjoin%
\definecolor{currentfill}{rgb}{0.882353,0.341176,0.349020}%
\pgfsetfillcolor{currentfill}%
\pgfsetlinewidth{1.003750pt}%
\definecolor{currentstroke}{rgb}{0.882353,0.341176,0.349020}%
\pgfsetstrokecolor{currentstroke}%
\pgfsetdash{}{0pt}%
\pgfsys@defobject{currentmarker}{\pgfqpoint{-0.027778in}{-0.000000in}}{\pgfqpoint{0.027778in}{0.000000in}}{%
\pgfpathmoveto{\pgfqpoint{0.027778in}{-0.000000in}}%
\pgfpathlineto{\pgfqpoint{-0.027778in}{0.000000in}}%
\pgfusepath{stroke,fill}%
}%
\begin{pgfscope}%
\pgfsys@transformshift{0.747091in}{1.339505in}%
\pgfsys@useobject{currentmarker}{}%
\end{pgfscope}%
\begin{pgfscope}%
\pgfsys@transformshift{0.839070in}{1.093538in}%
\pgfsys@useobject{currentmarker}{}%
\end{pgfscope}%
\begin{pgfscope}%
\pgfsys@transformshift{1.115008in}{0.764217in}%
\pgfsys@useobject{currentmarker}{}%
\end{pgfscope}%
\begin{pgfscope}%
\pgfsys@transformshift{1.574904in}{0.531776in}%
\pgfsys@useobject{currentmarker}{}%
\end{pgfscope}%
\begin{pgfscope}%
\pgfsys@transformshift{2.034800in}{0.413808in}%
\pgfsys@useobject{currentmarker}{}%
\end{pgfscope}%
\end{pgfscope}%
\begin{pgfscope}%
\pgfpathrectangle{\pgfqpoint{0.681786in}{0.358869in}}{\pgfqpoint{1.436716in}{1.208651in}}%
\pgfusepath{clip}%
\pgfsetbuttcap%
\pgfsetroundjoin%
\definecolor{currentfill}{rgb}{0.882353,0.341176,0.349020}%
\pgfsetfillcolor{currentfill}%
\pgfsetlinewidth{1.003750pt}%
\definecolor{currentstroke}{rgb}{0.882353,0.341176,0.349020}%
\pgfsetstrokecolor{currentstroke}%
\pgfsetdash{}{0pt}%
\pgfsys@defobject{currentmarker}{\pgfqpoint{-0.027778in}{-0.000000in}}{\pgfqpoint{0.027778in}{0.000000in}}{%
\pgfpathmoveto{\pgfqpoint{0.027778in}{-0.000000in}}%
\pgfpathlineto{\pgfqpoint{-0.027778in}{0.000000in}}%
\pgfusepath{stroke,fill}%
}%
\begin{pgfscope}%
\pgfsys@transformshift{0.747091in}{1.367152in}%
\pgfsys@useobject{currentmarker}{}%
\end{pgfscope}%
\begin{pgfscope}%
\pgfsys@transformshift{0.839070in}{1.129944in}%
\pgfsys@useobject{currentmarker}{}%
\end{pgfscope}%
\begin{pgfscope}%
\pgfsys@transformshift{1.115008in}{0.793166in}%
\pgfsys@useobject{currentmarker}{}%
\end{pgfscope}%
\begin{pgfscope}%
\pgfsys@transformshift{1.574904in}{0.574120in}%
\pgfsys@useobject{currentmarker}{}%
\end{pgfscope}%
\begin{pgfscope}%
\pgfsys@transformshift{2.034800in}{0.436661in}%
\pgfsys@useobject{currentmarker}{}%
\end{pgfscope}%
\end{pgfscope}%
\begin{pgfscope}%
\pgfpathrectangle{\pgfqpoint{0.681786in}{0.358869in}}{\pgfqpoint{1.436716in}{1.208651in}}%
\pgfusepath{clip}%
\pgfsetbuttcap%
\pgfsetroundjoin%
\pgfsetlinewidth{1.003750pt}%
\definecolor{currentstroke}{rgb}{0.305882,0.474510,0.654902}%
\pgfsetstrokecolor{currentstroke}%
\pgfsetdash{}{0pt}%
\pgfpathmoveto{\pgfqpoint{0.747091in}{1.232340in}}%
\pgfpathlineto{\pgfqpoint{0.747091in}{1.261708in}}%
\pgfusepath{stroke}%
\end{pgfscope}%
\begin{pgfscope}%
\pgfpathrectangle{\pgfqpoint{0.681786in}{0.358869in}}{\pgfqpoint{1.436716in}{1.208651in}}%
\pgfusepath{clip}%
\pgfsetbuttcap%
\pgfsetroundjoin%
\pgfsetlinewidth{1.003750pt}%
\definecolor{currentstroke}{rgb}{0.305882,0.474510,0.654902}%
\pgfsetstrokecolor{currentstroke}%
\pgfsetdash{}{0pt}%
\pgfpathmoveto{\pgfqpoint{0.839070in}{0.972868in}}%
\pgfpathlineto{\pgfqpoint{0.839070in}{1.004983in}}%
\pgfusepath{stroke}%
\end{pgfscope}%
\begin{pgfscope}%
\pgfpathrectangle{\pgfqpoint{0.681786in}{0.358869in}}{\pgfqpoint{1.436716in}{1.208651in}}%
\pgfusepath{clip}%
\pgfsetbuttcap%
\pgfsetroundjoin%
\pgfsetlinewidth{1.003750pt}%
\definecolor{currentstroke}{rgb}{0.305882,0.474510,0.654902}%
\pgfsetstrokecolor{currentstroke}%
\pgfsetdash{}{0pt}%
\pgfpathmoveto{\pgfqpoint{1.115008in}{0.720257in}}%
\pgfpathlineto{\pgfqpoint{1.115008in}{0.747556in}}%
\pgfusepath{stroke}%
\end{pgfscope}%
\begin{pgfscope}%
\pgfpathrectangle{\pgfqpoint{0.681786in}{0.358869in}}{\pgfqpoint{1.436716in}{1.208651in}}%
\pgfusepath{clip}%
\pgfsetbuttcap%
\pgfsetroundjoin%
\pgfsetlinewidth{1.003750pt}%
\definecolor{currentstroke}{rgb}{0.305882,0.474510,0.654902}%
\pgfsetstrokecolor{currentstroke}%
\pgfsetdash{}{0pt}%
\pgfpathmoveto{\pgfqpoint{1.574904in}{0.483367in}}%
\pgfpathlineto{\pgfqpoint{1.574904in}{0.538375in}}%
\pgfusepath{stroke}%
\end{pgfscope}%
\begin{pgfscope}%
\pgfpathrectangle{\pgfqpoint{0.681786in}{0.358869in}}{\pgfqpoint{1.436716in}{1.208651in}}%
\pgfusepath{clip}%
\pgfsetbuttcap%
\pgfsetroundjoin%
\pgfsetlinewidth{1.003750pt}%
\definecolor{currentstroke}{rgb}{0.305882,0.474510,0.654902}%
\pgfsetstrokecolor{currentstroke}%
\pgfsetdash{}{0pt}%
\pgfpathmoveto{\pgfqpoint{1.574904in}{0.502207in}}%
\pgfpathlineto{\pgfqpoint{1.574904in}{0.528663in}}%
\pgfusepath{stroke}%
\end{pgfscope}%
\begin{pgfscope}%
\pgfpathrectangle{\pgfqpoint{0.681786in}{0.358869in}}{\pgfqpoint{1.436716in}{1.208651in}}%
\pgfusepath{clip}%
\pgfsetbuttcap%
\pgfsetroundjoin%
\pgfsetlinewidth{1.003750pt}%
\definecolor{currentstroke}{rgb}{0.305882,0.474510,0.654902}%
\pgfsetstrokecolor{currentstroke}%
\pgfsetdash{}{0pt}%
\pgfpathmoveto{\pgfqpoint{2.034800in}{0.447932in}}%
\pgfpathlineto{\pgfqpoint{2.034800in}{0.479268in}}%
\pgfusepath{stroke}%
\end{pgfscope}%
\begin{pgfscope}%
\pgfpathrectangle{\pgfqpoint{0.681786in}{0.358869in}}{\pgfqpoint{1.436716in}{1.208651in}}%
\pgfusepath{clip}%
\pgfsetbuttcap%
\pgfsetroundjoin%
\definecolor{currentfill}{rgb}{0.305882,0.474510,0.654902}%
\pgfsetfillcolor{currentfill}%
\pgfsetlinewidth{1.003750pt}%
\definecolor{currentstroke}{rgb}{0.305882,0.474510,0.654902}%
\pgfsetstrokecolor{currentstroke}%
\pgfsetdash{}{0pt}%
\pgfsys@defobject{currentmarker}{\pgfqpoint{-0.027778in}{-0.000000in}}{\pgfqpoint{0.027778in}{0.000000in}}{%
\pgfpathmoveto{\pgfqpoint{0.027778in}{-0.000000in}}%
\pgfpathlineto{\pgfqpoint{-0.027778in}{0.000000in}}%
\pgfusepath{stroke,fill}%
}%
\begin{pgfscope}%
\pgfsys@transformshift{0.747091in}{1.232340in}%
\pgfsys@useobject{currentmarker}{}%
\end{pgfscope}%
\begin{pgfscope}%
\pgfsys@transformshift{0.839070in}{0.972868in}%
\pgfsys@useobject{currentmarker}{}%
\end{pgfscope}%
\begin{pgfscope}%
\pgfsys@transformshift{1.115008in}{0.720257in}%
\pgfsys@useobject{currentmarker}{}%
\end{pgfscope}%
\begin{pgfscope}%
\pgfsys@transformshift{1.574904in}{0.483367in}%
\pgfsys@useobject{currentmarker}{}%
\end{pgfscope}%
\begin{pgfscope}%
\pgfsys@transformshift{1.574904in}{0.502207in}%
\pgfsys@useobject{currentmarker}{}%
\end{pgfscope}%
\begin{pgfscope}%
\pgfsys@transformshift{2.034800in}{0.447932in}%
\pgfsys@useobject{currentmarker}{}%
\end{pgfscope}%
\end{pgfscope}%
\begin{pgfscope}%
\pgfpathrectangle{\pgfqpoint{0.681786in}{0.358869in}}{\pgfqpoint{1.436716in}{1.208651in}}%
\pgfusepath{clip}%
\pgfsetbuttcap%
\pgfsetroundjoin%
\definecolor{currentfill}{rgb}{0.305882,0.474510,0.654902}%
\pgfsetfillcolor{currentfill}%
\pgfsetlinewidth{1.003750pt}%
\definecolor{currentstroke}{rgb}{0.305882,0.474510,0.654902}%
\pgfsetstrokecolor{currentstroke}%
\pgfsetdash{}{0pt}%
\pgfsys@defobject{currentmarker}{\pgfqpoint{-0.027778in}{-0.000000in}}{\pgfqpoint{0.027778in}{0.000000in}}{%
\pgfpathmoveto{\pgfqpoint{0.027778in}{-0.000000in}}%
\pgfpathlineto{\pgfqpoint{-0.027778in}{0.000000in}}%
\pgfusepath{stroke,fill}%
}%
\begin{pgfscope}%
\pgfsys@transformshift{0.747091in}{1.261708in}%
\pgfsys@useobject{currentmarker}{}%
\end{pgfscope}%
\begin{pgfscope}%
\pgfsys@transformshift{0.839070in}{1.004983in}%
\pgfsys@useobject{currentmarker}{}%
\end{pgfscope}%
\begin{pgfscope}%
\pgfsys@transformshift{1.115008in}{0.747556in}%
\pgfsys@useobject{currentmarker}{}%
\end{pgfscope}%
\begin{pgfscope}%
\pgfsys@transformshift{1.574904in}{0.538375in}%
\pgfsys@useobject{currentmarker}{}%
\end{pgfscope}%
\begin{pgfscope}%
\pgfsys@transformshift{1.574904in}{0.528663in}%
\pgfsys@useobject{currentmarker}{}%
\end{pgfscope}%
\begin{pgfscope}%
\pgfsys@transformshift{2.034800in}{0.479268in}%
\pgfsys@useobject{currentmarker}{}%
\end{pgfscope}%
\end{pgfscope}%
\begin{pgfscope}%
\pgfpathrectangle{\pgfqpoint{0.681786in}{0.358869in}}{\pgfqpoint{1.436716in}{1.208651in}}%
\pgfusepath{clip}%
\pgfsetroundcap%
\pgfsetroundjoin%
\pgfsetlinewidth{1.204500pt}%
\definecolor{currentstroke}{rgb}{0.882353,0.341176,0.349020}%
\pgfsetstrokecolor{currentstroke}%
\pgfsetdash{}{0pt}%
\pgfpathmoveto{\pgfqpoint{0.765487in}{1.497702in}}%
\pgfpathlineto{\pgfqpoint{0.857466in}{1.345277in}}%
\pgfpathlineto{\pgfqpoint{1.133404in}{1.052300in}}%
\pgfpathlineto{\pgfqpoint{1.593300in}{0.744555in}}%
\pgfpathlineto{\pgfqpoint{2.053196in}{0.544748in}}%
\pgfusepath{stroke}%
\end{pgfscope}%
\begin{pgfscope}%
\pgfpathrectangle{\pgfqpoint{0.681786in}{0.358869in}}{\pgfqpoint{1.436716in}{1.208651in}}%
\pgfusepath{clip}%
\pgfsetbuttcap%
\pgfsetroundjoin%
\pgfsetlinewidth{1.204500pt}%
\definecolor{currentstroke}{rgb}{0.882353,0.341176,0.349020}%
\pgfsetstrokecolor{currentstroke}%
\pgfsetdash{{1.200000pt}{1.980000pt}}{0.000000pt}%
\pgfpathmoveto{\pgfqpoint{0.765487in}{1.411063in}}%
\pgfpathlineto{\pgfqpoint{0.857466in}{1.224242in}}%
\pgfpathlineto{\pgfqpoint{1.133404in}{0.956756in}}%
\pgfpathlineto{\pgfqpoint{1.593300in}{0.703889in}}%
\pgfpathlineto{\pgfqpoint{2.053196in}{0.526195in}}%
\pgfusepath{stroke}%
\end{pgfscope}%
\begin{pgfscope}%
\pgfpathrectangle{\pgfqpoint{0.681786in}{0.358869in}}{\pgfqpoint{1.436716in}{1.208651in}}%
\pgfusepath{clip}%
\pgfsetbuttcap%
\pgfsetroundjoin%
\pgfsetlinewidth{1.204500pt}%
\definecolor{currentstroke}{rgb}{0.949020,0.556863,0.168627}%
\pgfsetstrokecolor{currentstroke}%
\pgfsetdash{{4.440000pt}{1.920000pt}}{0.000000pt}%
\pgfpathmoveto{\pgfqpoint{0.747091in}{1.353328in}}%
\pgfpathlineto{\pgfqpoint{0.839070in}{1.111741in}}%
\pgfpathlineto{\pgfqpoint{1.115008in}{0.778691in}}%
\pgfpathlineto{\pgfqpoint{1.574904in}{0.552948in}}%
\pgfpathlineto{\pgfqpoint{2.034800in}{0.425234in}}%
\pgfusepath{stroke}%
\end{pgfscope}%
\begin{pgfscope}%
\pgfpathrectangle{\pgfqpoint{0.681786in}{0.358869in}}{\pgfqpoint{1.436716in}{1.208651in}}%
\pgfusepath{clip}%
\pgfsetroundcap%
\pgfsetroundjoin%
\pgfsetlinewidth{1.204500pt}%
\definecolor{currentstroke}{rgb}{0.305882,0.474510,0.654902}%
\pgfsetstrokecolor{currentstroke}%
\pgfsetdash{}{0pt}%
\pgfpathmoveto{\pgfqpoint{0.747091in}{1.247024in}}%
\pgfpathlineto{\pgfqpoint{0.839070in}{0.988925in}}%
\pgfpathlineto{\pgfqpoint{1.115008in}{0.733906in}}%
\pgfpathlineto{\pgfqpoint{1.574904in}{0.510871in}}%
\pgfpathlineto{\pgfqpoint{1.574904in}{0.515435in}}%
\pgfpathlineto{\pgfqpoint{2.034800in}{0.463600in}}%
\pgfusepath{stroke}%
\end{pgfscope}%
\begin{pgfscope}%
\pgfsetrectcap%
\pgfsetmiterjoin%
\pgfsetlinewidth{1.003750pt}%
\definecolor{currentstroke}{rgb}{0.800000,0.800000,0.800000}%
\pgfsetstrokecolor{currentstroke}%
\pgfsetdash{}{0pt}%
\pgfpathmoveto{\pgfqpoint{0.681786in}{0.358869in}}%
\pgfpathlineto{\pgfqpoint{0.681786in}{1.567520in}}%
\pgfusepath{stroke}%
\end{pgfscope}%
\begin{pgfscope}%
\pgfsetrectcap%
\pgfsetmiterjoin%
\pgfsetlinewidth{1.003750pt}%
\definecolor{currentstroke}{rgb}{0.800000,0.800000,0.800000}%
\pgfsetstrokecolor{currentstroke}%
\pgfsetdash{}{0pt}%
\pgfpathmoveto{\pgfqpoint{2.118501in}{0.358869in}}%
\pgfpathlineto{\pgfqpoint{2.118501in}{1.567520in}}%
\pgfusepath{stroke}%
\end{pgfscope}%
\begin{pgfscope}%
\pgfsetrectcap%
\pgfsetmiterjoin%
\pgfsetlinewidth{1.003750pt}%
\definecolor{currentstroke}{rgb}{0.800000,0.800000,0.800000}%
\pgfsetstrokecolor{currentstroke}%
\pgfsetdash{}{0pt}%
\pgfpathmoveto{\pgfqpoint{0.681786in}{0.358869in}}%
\pgfpathlineto{\pgfqpoint{2.118501in}{0.358869in}}%
\pgfusepath{stroke}%
\end{pgfscope}%
\begin{pgfscope}%
\pgfsetrectcap%
\pgfsetmiterjoin%
\pgfsetlinewidth{1.003750pt}%
\definecolor{currentstroke}{rgb}{0.800000,0.800000,0.800000}%
\pgfsetstrokecolor{currentstroke}%
\pgfsetdash{}{0pt}%
\pgfpathmoveto{\pgfqpoint{0.681786in}{1.567520in}}%
\pgfpathlineto{\pgfqpoint{2.118501in}{1.567520in}}%
\pgfusepath{stroke}%
\end{pgfscope}%
\begin{pgfscope}%
\definecolor{textcolor}{rgb}{0.150000,0.150000,0.150000}%
\pgfsetstrokecolor{textcolor}%
\pgfsetfillcolor{textcolor}%
\pgftext[x=0.423177in,y=0.213831in,,]{\color{textcolor}{\rmfamily\fontsize{8.832000}{10.598400}\selectfont\catcode`\^=\active\def^{\ifmmode\sp\else\^{}\fi}\catcode`\%=\active\def
\end{pgfscope}%
\begin{pgfscope}%
\pgfsetbuttcap%
\pgfsetmiterjoin%
\definecolor{currentfill}{rgb}{1.000000,1.000000,1.000000}%
\pgfsetfillcolor{currentfill}%
\pgfsetfillopacity{0.800000}%
\pgfsetlinewidth{0.803000pt}%
\definecolor{currentstroke}{rgb}{0.800000,0.800000,0.800000}%
\pgfsetstrokecolor{currentstroke}%
\pgfsetstrokeopacity{0.800000}%
\pgfsetdash{}{0pt}%
\pgfpathmoveto{\pgfqpoint{1.082035in}{1.011596in}}%
\pgfpathlineto{\pgfqpoint{2.060168in}{1.011596in}}%
\pgfpathquadraticcurveto{\pgfqpoint{2.076834in}{1.011596in}}{\pgfqpoint{2.076834in}{1.028263in}}%
\pgfpathlineto{\pgfqpoint{2.076834in}{1.509187in}}%
\pgfpathquadraticcurveto{\pgfqpoint{2.076834in}{1.525853in}}{\pgfqpoint{2.060168in}{1.525853in}}%
\pgfpathlineto{\pgfqpoint{1.082035in}{1.525853in}}%
\pgfpathquadraticcurveto{\pgfqpoint{1.065368in}{1.525853in}}{\pgfqpoint{1.065368in}{1.509187in}}%
\pgfpathlineto{\pgfqpoint{1.065368in}{1.028263in}}%
\pgfpathquadraticcurveto{\pgfqpoint{1.065368in}{1.011596in}}{\pgfqpoint{1.082035in}{1.011596in}}%
\pgfpathlineto{\pgfqpoint{1.082035in}{1.011596in}}%
\pgfpathclose%
\pgfusepath{stroke,fill}%
\end{pgfscope}%
\begin{pgfscope}%
\pgfsetbuttcap%
\pgfsetroundjoin%
\pgfsetlinewidth{1.003750pt}%
\definecolor{currentstroke}{rgb}{0.882353,0.341176,0.349020}%
\pgfsetstrokecolor{currentstroke}%
\pgfsetdash{}{0pt}%
\pgfpathmoveto{\pgfqpoint{1.182035in}{1.416706in}}%
\pgfpathlineto{\pgfqpoint{1.182035in}{1.500039in}}%
\pgfusepath{stroke}%
\end{pgfscope}%
\begin{pgfscope}%
\pgfsetbuttcap%
\pgfsetroundjoin%
\definecolor{currentfill}{rgb}{0.882353,0.341176,0.349020}%
\pgfsetfillcolor{currentfill}%
\pgfsetlinewidth{1.003750pt}%
\definecolor{currentstroke}{rgb}{0.882353,0.341176,0.349020}%
\pgfsetstrokecolor{currentstroke}%
\pgfsetdash{}{0pt}%
\pgfsys@defobject{currentmarker}{\pgfqpoint{-0.027778in}{-0.000000in}}{\pgfqpoint{0.027778in}{0.000000in}}{%
\pgfpathmoveto{\pgfqpoint{0.027778in}{-0.000000in}}%
\pgfpathlineto{\pgfqpoint{-0.027778in}{0.000000in}}%
\pgfusepath{stroke,fill}%
}%
\begin{pgfscope}%
\pgfsys@transformshift{1.182035in}{1.416706in}%
\pgfsys@useobject{currentmarker}{}%
\end{pgfscope}%
\end{pgfscope}%
\begin{pgfscope}%
\pgfsetbuttcap%
\pgfsetroundjoin%
\definecolor{currentfill}{rgb}{0.882353,0.341176,0.349020}%
\pgfsetfillcolor{currentfill}%
\pgfsetlinewidth{1.003750pt}%
\definecolor{currentstroke}{rgb}{0.882353,0.341176,0.349020}%
\pgfsetstrokecolor{currentstroke}%
\pgfsetdash{}{0pt}%
\pgfsys@defobject{currentmarker}{\pgfqpoint{-0.027778in}{-0.000000in}}{\pgfqpoint{0.027778in}{0.000000in}}{%
\pgfpathmoveto{\pgfqpoint{0.027778in}{-0.000000in}}%
\pgfpathlineto{\pgfqpoint{-0.027778in}{0.000000in}}%
\pgfusepath{stroke,fill}%
}%
\begin{pgfscope}%
\pgfsys@transformshift{1.182035in}{1.500039in}%
\pgfsys@useobject{currentmarker}{}%
\end{pgfscope}%
\end{pgfscope}%
\begin{pgfscope}%
\pgfsetroundcap%
\pgfsetroundjoin%
\pgfsetlinewidth{1.204500pt}%
\definecolor{currentstroke}{rgb}{0.882353,0.341176,0.349020}%
\pgfsetstrokecolor{currentstroke}%
\pgfsetdash{}{0pt}%
\pgfpathmoveto{\pgfqpoint{1.098701in}{1.458373in}}%
\pgfpathlineto{\pgfqpoint{1.265368in}{1.458373in}}%
\pgfusepath{stroke}%
\end{pgfscope}%
\begin{pgfscope}%
\definecolor{textcolor}{rgb}{0.150000,0.150000,0.150000}%
\pgfsetstrokecolor{textcolor}%
\pgfsetfillcolor{textcolor}%
\pgftext[x=1.332035in,y=1.429206in,left,base]{\color{textcolor}{\rmfamily\fontsize{6.000000}{7.200000}\selectfont\catcode`\^=\active\def^{\ifmmode\sp\else\^{}\fi}\catcode`\%=\active\def
\end{pgfscope}%
\begin{pgfscope}%
\pgfsetbuttcap%
\pgfsetroundjoin%
\pgfsetlinewidth{1.003750pt}%
\definecolor{currentstroke}{rgb}{0.882353,0.341176,0.349020}%
\pgfsetstrokecolor{currentstroke}%
\pgfsetdash{}{0pt}%
\pgfpathmoveto{\pgfqpoint{1.182035in}{1.294392in}}%
\pgfpathlineto{\pgfqpoint{1.182035in}{1.377725in}}%
\pgfusepath{stroke}%
\end{pgfscope}%
\begin{pgfscope}%
\pgfsetbuttcap%
\pgfsetroundjoin%
\definecolor{currentfill}{rgb}{0.882353,0.341176,0.349020}%
\pgfsetfillcolor{currentfill}%
\pgfsetlinewidth{1.003750pt}%
\definecolor{currentstroke}{rgb}{0.882353,0.341176,0.349020}%
\pgfsetstrokecolor{currentstroke}%
\pgfsetdash{}{0pt}%
\pgfsys@defobject{currentmarker}{\pgfqpoint{-0.027778in}{-0.000000in}}{\pgfqpoint{0.027778in}{0.000000in}}{%
\pgfpathmoveto{\pgfqpoint{0.027778in}{-0.000000in}}%
\pgfpathlineto{\pgfqpoint{-0.027778in}{0.000000in}}%
\pgfusepath{stroke,fill}%
}%
\begin{pgfscope}%
\pgfsys@transformshift{1.182035in}{1.294392in}%
\pgfsys@useobject{currentmarker}{}%
\end{pgfscope}%
\end{pgfscope}%
\begin{pgfscope}%
\pgfsetbuttcap%
\pgfsetroundjoin%
\definecolor{currentfill}{rgb}{0.882353,0.341176,0.349020}%
\pgfsetfillcolor{currentfill}%
\pgfsetlinewidth{1.003750pt}%
\definecolor{currentstroke}{rgb}{0.882353,0.341176,0.349020}%
\pgfsetstrokecolor{currentstroke}%
\pgfsetdash{}{0pt}%
\pgfsys@defobject{currentmarker}{\pgfqpoint{-0.027778in}{-0.000000in}}{\pgfqpoint{0.027778in}{0.000000in}}{%
\pgfpathmoveto{\pgfqpoint{0.027778in}{-0.000000in}}%
\pgfpathlineto{\pgfqpoint{-0.027778in}{0.000000in}}%
\pgfusepath{stroke,fill}%
}%
\begin{pgfscope}%
\pgfsys@transformshift{1.182035in}{1.377725in}%
\pgfsys@useobject{currentmarker}{}%
\end{pgfscope}%
\end{pgfscope}%
\begin{pgfscope}%
\pgfsetbuttcap%
\pgfsetroundjoin%
\pgfsetlinewidth{1.204500pt}%
\definecolor{currentstroke}{rgb}{0.882353,0.341176,0.349020}%
\pgfsetstrokecolor{currentstroke}%
\pgfsetdash{{1.200000pt}{1.980000pt}}{0.000000pt}%
\pgfpathmoveto{\pgfqpoint{1.098701in}{1.336058in}}%
\pgfpathlineto{\pgfqpoint{1.265368in}{1.336058in}}%
\pgfusepath{stroke}%
\end{pgfscope}%
\begin{pgfscope}%
\definecolor{textcolor}{rgb}{0.150000,0.150000,0.150000}%
\pgfsetstrokecolor{textcolor}%
\pgfsetfillcolor{textcolor}%
\pgftext[x=1.332035in,y=1.306892in,left,base]{\color{textcolor}{\rmfamily\fontsize{6.000000}{7.200000}\selectfont\catcode`\^=\active\def^{\ifmmode\sp\else\^{}\fi}\catcode`\%=\active\def
\end{pgfscope}%
\begin{pgfscope}%
\pgfsetbuttcap%
\pgfsetroundjoin%
\pgfsetlinewidth{1.003750pt}%
\definecolor{currentstroke}{rgb}{0.882353,0.341176,0.349020}%
\pgfsetstrokecolor{currentstroke}%
\pgfsetdash{}{0pt}%
\pgfpathmoveto{\pgfqpoint{1.182035in}{1.172077in}}%
\pgfpathlineto{\pgfqpoint{1.182035in}{1.255411in}}%
\pgfusepath{stroke}%
\end{pgfscope}%
\begin{pgfscope}%
\pgfsetbuttcap%
\pgfsetroundjoin%
\definecolor{currentfill}{rgb}{0.882353,0.341176,0.349020}%
\pgfsetfillcolor{currentfill}%
\pgfsetlinewidth{1.003750pt}%
\definecolor{currentstroke}{rgb}{0.882353,0.341176,0.349020}%
\pgfsetstrokecolor{currentstroke}%
\pgfsetdash{}{0pt}%
\pgfsys@defobject{currentmarker}{\pgfqpoint{-0.027778in}{-0.000000in}}{\pgfqpoint{0.027778in}{0.000000in}}{%
\pgfpathmoveto{\pgfqpoint{0.027778in}{-0.000000in}}%
\pgfpathlineto{\pgfqpoint{-0.027778in}{0.000000in}}%
\pgfusepath{stroke,fill}%
}%
\begin{pgfscope}%
\pgfsys@transformshift{1.182035in}{1.172077in}%
\pgfsys@useobject{currentmarker}{}%
\end{pgfscope}%
\end{pgfscope}%
\begin{pgfscope}%
\pgfsetbuttcap%
\pgfsetroundjoin%
\definecolor{currentfill}{rgb}{0.882353,0.341176,0.349020}%
\pgfsetfillcolor{currentfill}%
\pgfsetlinewidth{1.003750pt}%
\definecolor{currentstroke}{rgb}{0.882353,0.341176,0.349020}%
\pgfsetstrokecolor{currentstroke}%
\pgfsetdash{}{0pt}%
\pgfsys@defobject{currentmarker}{\pgfqpoint{-0.027778in}{-0.000000in}}{\pgfqpoint{0.027778in}{0.000000in}}{%
\pgfpathmoveto{\pgfqpoint{0.027778in}{-0.000000in}}%
\pgfpathlineto{\pgfqpoint{-0.027778in}{0.000000in}}%
\pgfusepath{stroke,fill}%
}%
\begin{pgfscope}%
\pgfsys@transformshift{1.182035in}{1.255411in}%
\pgfsys@useobject{currentmarker}{}%
\end{pgfscope}%
\end{pgfscope}%
\begin{pgfscope}%
\pgfsetbuttcap%
\pgfsetroundjoin%
\pgfsetlinewidth{1.204500pt}%
\definecolor{currentstroke}{rgb}{0.949020,0.556863,0.168627}%
\pgfsetstrokecolor{currentstroke}%
\pgfsetdash{{4.440000pt}{1.920000pt}}{0.000000pt}%
\pgfpathmoveto{\pgfqpoint{1.098701in}{1.213744in}}%
\pgfpathlineto{\pgfqpoint{1.265368in}{1.213744in}}%
\pgfusepath{stroke}%
\end{pgfscope}%
\begin{pgfscope}%
\definecolor{textcolor}{rgb}{0.150000,0.150000,0.150000}%
\pgfsetstrokecolor{textcolor}%
\pgfsetfillcolor{textcolor}%
\pgftext[x=1.332035in,y=1.184577in,left,base]{\color{textcolor}{\rmfamily\fontsize{6.000000}{7.200000}\selectfont\catcode`\^=\active\def^{\ifmmode\sp\else\^{}\fi}\catcode`\%=\active\def
\end{pgfscope}%
\begin{pgfscope}%
\pgfsetbuttcap%
\pgfsetroundjoin%
\pgfsetlinewidth{1.003750pt}%
\definecolor{currentstroke}{rgb}{0.305882,0.474510,0.654902}%
\pgfsetstrokecolor{currentstroke}%
\pgfsetdash{}{0pt}%
\pgfpathmoveto{\pgfqpoint{1.182035in}{1.049763in}}%
\pgfpathlineto{\pgfqpoint{1.182035in}{1.133097in}}%
\pgfusepath{stroke}%
\end{pgfscope}%
\begin{pgfscope}%
\pgfsetbuttcap%
\pgfsetroundjoin%
\definecolor{currentfill}{rgb}{0.305882,0.474510,0.654902}%
\pgfsetfillcolor{currentfill}%
\pgfsetlinewidth{1.003750pt}%
\definecolor{currentstroke}{rgb}{0.305882,0.474510,0.654902}%
\pgfsetstrokecolor{currentstroke}%
\pgfsetdash{}{0pt}%
\pgfsys@defobject{currentmarker}{\pgfqpoint{-0.027778in}{-0.000000in}}{\pgfqpoint{0.027778in}{0.000000in}}{%
\pgfpathmoveto{\pgfqpoint{0.027778in}{-0.000000in}}%
\pgfpathlineto{\pgfqpoint{-0.027778in}{0.000000in}}%
\pgfusepath{stroke,fill}%
}%
\begin{pgfscope}%
\pgfsys@transformshift{1.182035in}{1.049763in}%
\pgfsys@useobject{currentmarker}{}%
\end{pgfscope}%
\end{pgfscope}%
\begin{pgfscope}%
\pgfsetbuttcap%
\pgfsetroundjoin%
\definecolor{currentfill}{rgb}{0.305882,0.474510,0.654902}%
\pgfsetfillcolor{currentfill}%
\pgfsetlinewidth{1.003750pt}%
\definecolor{currentstroke}{rgb}{0.305882,0.474510,0.654902}%
\pgfsetstrokecolor{currentstroke}%
\pgfsetdash{}{0pt}%
\pgfsys@defobject{currentmarker}{\pgfqpoint{-0.027778in}{-0.000000in}}{\pgfqpoint{0.027778in}{0.000000in}}{%
\pgfpathmoveto{\pgfqpoint{0.027778in}{-0.000000in}}%
\pgfpathlineto{\pgfqpoint{-0.027778in}{0.000000in}}%
\pgfusepath{stroke,fill}%
}%
\begin{pgfscope}%
\pgfsys@transformshift{1.182035in}{1.133097in}%
\pgfsys@useobject{currentmarker}{}%
\end{pgfscope}%
\end{pgfscope}%
\begin{pgfscope}%
\pgfsetroundcap%
\pgfsetroundjoin%
\pgfsetlinewidth{1.204500pt}%
\definecolor{currentstroke}{rgb}{0.305882,0.474510,0.654902}%
\pgfsetstrokecolor{currentstroke}%
\pgfsetdash{}{0pt}%
\pgfpathmoveto{\pgfqpoint{1.098701in}{1.091430in}}%
\pgfpathlineto{\pgfqpoint{1.265368in}{1.091430in}}%
\pgfusepath{stroke}%
\end{pgfscope}%
\begin{pgfscope}%
\definecolor{textcolor}{rgb}{0.150000,0.150000,0.150000}%
\pgfsetstrokecolor{textcolor}%
\pgfsetfillcolor{textcolor}%
\pgftext[x=1.332035in,y=1.062263in,left,base]{\color{textcolor}{\rmfamily\fontsize{6.000000}{7.200000}\selectfont\catcode`\^=\active\def^{\ifmmode\sp\else\^{}\fi}\catcode`\%=\active\def
\end{pgfscope}%
\end{pgfpicture}%
\makeatother%
\endgroup%